\documentclass[10pt,twocolumn,letterpaper]{article}

\usepackage{iccv}

\usepackage{times}
\usepackage{epsfig}
\usepackage{graphicx}
\usepackage{amsmath}
\usepackage{amssymb}
\usepackage{booktabs}
\usepackage{xspace}
\usepackage{caption}
\usepackage{float}
\usepackage{xspace}
\usepackage{dblfloatfix}
\usepackage{inconsolata,enumitem}
\usepackage{xcolor}
\usepackage{multirow}

\usepackage[pagebackref=true,breaklinks=true,letterpaper=true,colorlinks,bookmarks=false]{hyperref}

\iccvfinalcopy 
\def\iccvPaperID{****}

\newcommand{\methodname}{Worldsheet\xspace}
\newcommand{\myparagraph}[1]{\vspace{0.04in}\noindent\textbf{#1.}\xspace}

\definecolor{demphcolor}{RGB}{110,110,110}
\newcommand{\demph}[1]{\textcolor{demphcolor}{#1}}

\newcounter{magicrownumbers}
\newcommand\rownumber{\stepcounter{magicrownumbers}\arabic{magicrownumbers}}

\begin{document}

\title{\methodname: Wrapping the World in a 3D Sheet\\for View Synthesis from a Single Image}

\author{Ronghang Hu$^1$ $\qquad$ Nikhila Ravi$^1$ $\qquad$ Alexander C. Berg$^1$ $\qquad$ Deepak Pathak$^2$ \\
$^1$Facebook AI Research (FAIR) $\qquad$ $^2$Carnegie Mellon University\\
}

\twocolumn[{\renewcommand\twocolumn[1][]{#1}\vspace{-1em}
\maketitle
\centering
\small
\vspace{-2.5em}
\includegraphics[width=\textwidth]{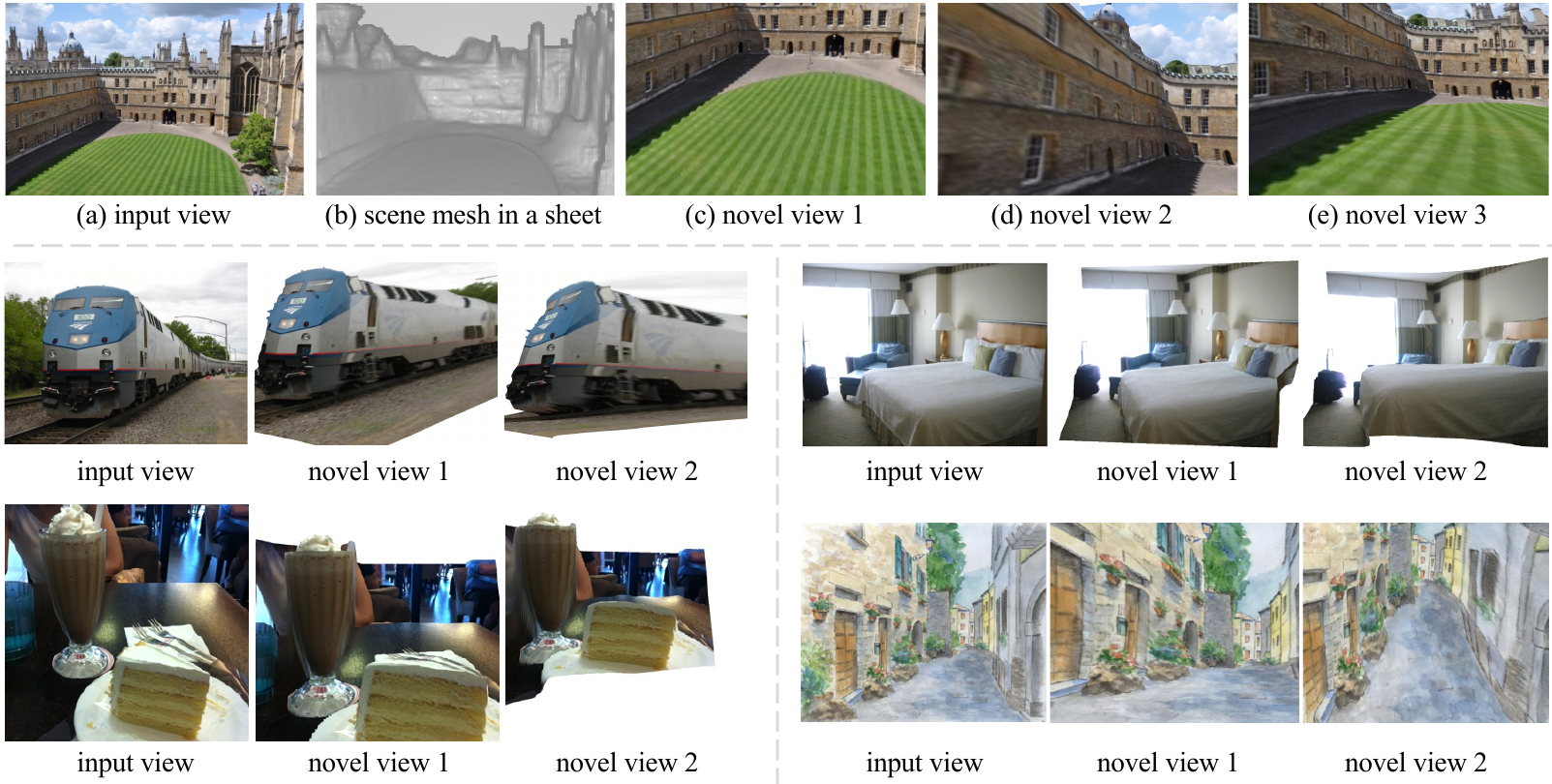} \\ \vspace{-0.5em}
\captionof{figure}{We synthesize novel views from large viewpoint changes given a single input RGB image (shown in a) by wrapping a mesh sheet (shown in b) onto the image, and rendering it from novel viewpoints (shown in c, d, e). Plausible novel views are generated for outdoor scenes, outdoor objects, indoor scenes, indoor objects and even paintings with high resolution ($960 \times 960$) input. Please see continuously synthesized views at \href{https://worldsheet.github.io}{\textcolor{black}{\texttt{worldsheet.github.io}}} (Image sources: \cite{img_oxford_new_college,hoiem2005automatic,lin2014microsoft,img_italian_alley}).}
\label{fig:vis_popup}
\vspace{1em}
}]

\begin{abstract}
\vspace{-1em}
We present Worldsheet, a method for novel view synthesis using just a single RGB image as input. The main insight is that simply shrink-wrapping a planar mesh sheet onto the input image, consistent with the learned intermediate depth, captures underlying geometry sufficient to generate photorealistic unseen views with large viewpoint changes. To operationalize this, we propose a novel differentiable texture sampler that allows our wrapped mesh sheet to be textured and rendered differentiably into an image from a target viewpoint. Our approach is category-agnostic, end-to-end trainable without using any 3D supervision, and requires a single image at test time. We also explore a simple extension by stacking multiple layers of Worldsheets to better handle occlusions. Worldsheet consistently outperforms prior state-of-the-art methods on single-image view synthesis across several datasets. Furthermore, this simple idea captures novel views surprisingly well on a wide range of high-resolution in-the-wild images, converting them into navigable 3D pop-ups. Video results and code are available at \url{https://worldsheet.github.io}.
\end{abstract}

\vspace{-1.5em}
\section{Introduction}
\label{sec:intro}
A 2D image is the projection of an underlying 3D world, but as humans, we have no trouble in understanding this structure and imagining how an image will look from other views. Consider the train shown in Figure~\ref{fig:vis_popup}, we can seamlessly predict other views from a single image based on the abstractions we have learned from past experience of seeing several trains, or similar shaped objects from different views. Enabling machines with such an ability to reason about 3D from a single image will bring trillions of still photos to life, with wide applications in virtual reality, animation, image editing, and robotics.

The goal of synthesizing novel views from 2D images has been pursued for decades, from early efforts relying completely on multi-view geometry~\cite{debevec1996modeling,zitnick2004high,seitz2006comparison}, to more recent learning based approaches~\cite{zhou2018stereo,sitzmann2019scene,srinivasan2019pushing,aliev2019neural,du2019project,mildenhall2020nerf,liu2020neural,riegler2020stable,li2020multi,wang2021ibrnet,wizadwongsa2021nex}. Over the years, significant progress has been made in this direction. However, despite impressive photorealistic output renderings, most of these previous approaches require multiple images or ground-truth depth at test time, which severely hinders their practicality. 
To compensate for the lack of multiple views or 3D models at test time, methods for single-image 3D rely on statistical learning from data. This line of work can be traced back to classic works of Hoiem~\etal~\cite{hoiem2005automatic}, followed by Saxena~\etal~\cite{saxena2008make3d}, that obtain `qualitative 3D' from a single image by fitting a collection of planes onto the image.

An ideal approach to general-purpose view synthesis should not only rely on a single image at test time, but also learn from easy-to-collect supervision signal during training. In the deep learning era, there is growing interest in end-to-end methods with intermediate 3D representations supervised by multiple images and no explicit 3D information during training. However, they are mostly applied to objects~\cite{kulkarni2015deep,tatarchenko2016multi,worrall2017interpretable,kanazawa2018learning,sitzmann2019scene}, and are either category-specific, restricted to synthetic scenes, or both. Recent works~\cite{chen2019monocular,wiles2020synsin,tucker2020single} address these issues by training with multiple views of real-world scenes, relying on point cloud or multiplane images as intermediate representations. However, multiplane images only perform well with relatively small viewpoint changes as each plane is at a constant depth; for point clouds, one needs to represent each point in a scene individually, making it inefficient to scale to high-resolution data or large viewpoint changes. In contrast, meshes can provide a sparser scene representation, \eg, two triangular mesh faces can theoretically represent the entire flat surface of a wall, making it ideal for single-image view synthesis. However, mesh recovery from single images has been studied mostly for object images and in a category-specific manner~\cite{blanz1999morphable,kanazawa2018learning,kulkarni2019canonical} and not for scenes.

\begin{figure*}
\vspace{-1.5em}
\centering
\includegraphics[width=0.95\linewidth]{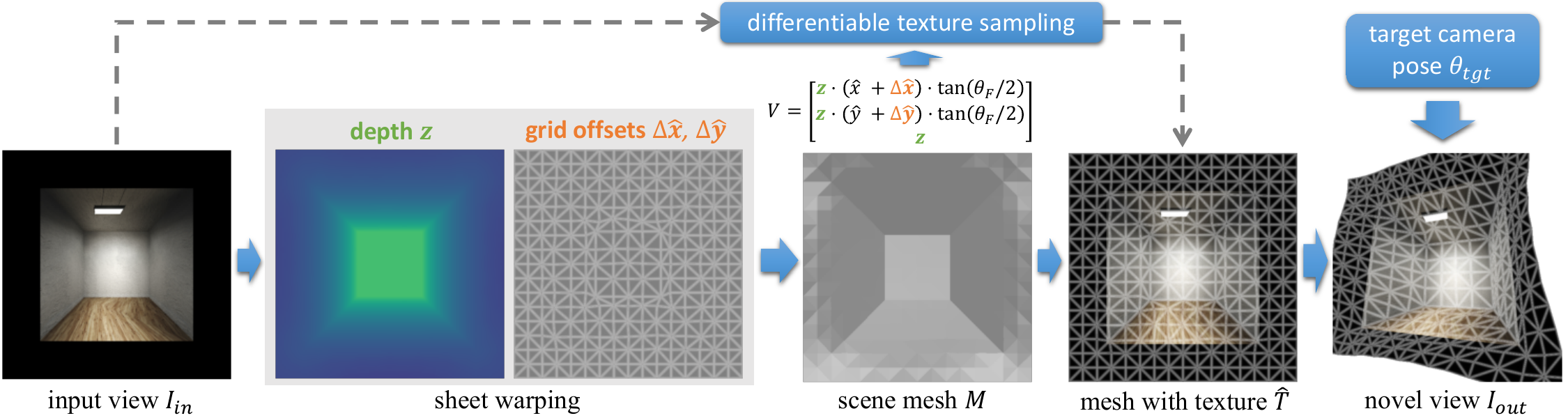}
\vspace{-0.5em}
\caption{An overview of our \textit{\methodname} approach. Given an input view $I_{in}$, we build a scene mesh by warping a $W_{m} \times H_{m}$ grid sheet onto the scene geometry via grid offset $(\Delta\hat{x}, \Delta\hat{y})$ and depth $z$ (Sec.~\ref{sec:method_mesh_pred}). Then, we sample the UV texture map $\hat{T}$ of the scene mesh differentiably (Sec.~\ref{sec:method_texture_sampling}) and render it from the target camera pose to output a novel view $I_{out}$. Our mesh warping is learned end-to-end using the losses on the novel view (Sec.~\ref{sec:method_analysis_by_synthesis}). We further apply an inpainting network over $I_{out}$ (not shown above; see Sec.~\ref{sec:method_inpainting}) to inpaint invisible regions and refine image details, outputting a refined novel view $I_{paint}$. We train with two views without any 3D or depth supervision and require just a single RGB image at test time.}
\label{fig:method}
\vspace{-1em}
\end{figure*}

In this paper, we present an end-to-end approach for novel view synthesis from \textit{a single image} of a scene via an intermediate \textit{mesh representation}. Unlike mesh reconstruction for objects of specific categories, generating meshes for a scene is challenging as there is no notion of mean or canonical shape to start from, or silhouette from segmentation for supervision. We circumvent this problem by wrapping a deformable mesh sheet over the 3D world -- much like wrapping a 2D tinfoil onto a 3D pan before baking! We name this
shrink-wrapped mesh \textit{\methodname}, a term borrowed from physics for the 2D manifold of high-dimensional strings. After generating this \methodname for a given view, novel views are obtained by moving the camera in 3D space (Figure~\ref{fig:method}), which allows us to train from just two views of a scene using only rendering losses \textit{without any 3D or depth supervision}.

To train our model end-to-end, both reconstruction of the mesh texture from input view and rendering from a novel camera view need to be differentiable. The latter is easily handled thanks to recent differentiable mesh renderers~\cite{kato2018neural,liu2019soft,ravi2020accelerating}. To address the former, we propose a differentiable texture sampler over projected 2D views, enabling gradient computation of the reconstructed texture map over the 3D mesh geometry. Furthermore, to better handle occlusions and depth discontinuities, we propose a simple extension by stacking multiple layers of Worldsheets onto the scene.

In summary, Worldsheet generates novel views by learning to predict scene geometry from a single image.
Although 3D mesh reconstruction via differentiable rendering is common for objects, to our best knowledge, this is the first work to show mesh recovery for \textit{scenes} just from multi-view supervision.
Our model consistently outperforms prior state-of-the-art by a significant margin on three benchmark datasets (Matterport \cite{chang2017matterport3d}, Replica \cite{straub2019replica}, and RealEstate10K \cite{zhou2018stereo}), and is applicable to very high-resolution images in-the-wild as shown in Figure~\ref{fig:vis_popup}.

\section{Related work}
\label{sec:related_work}

\myparagraph{Novel view synthesis from multiple images} Traditional novel view synthesis methods use multiple input views at test time~\cite{chen1995quicktime,levoy1996light,gortler1996lumigraph}, and are often based on different representations. Among recent works, Waechter \etal \cite{waechter2014let} build scene meshes with diffuse appearance. StereoMag~\cite{zhou2018stereo} proposes multiplane images (MPIs) from a stereo image pair as a layered scene representation. NPBG \cite{aliev2019neural} captures the scene as a point cloud with neural descriptors. NeRF \cite{mildenhall2020nerf} proposes a neural radiance field representation for scene appearance, and is followed by many extensions (see \cite{nerf_explosion_2020} for a summary). NSVF \cite{liu2020neural} adopts sparse voxel octrees as scene representations. FVS \cite{riegler2020free} and SVS \cite{riegler2020stable} blend multiple source images based on a geometric scaffold. Yoon \etal \cite{yoon2020novel} combine depth from both single and multiple views to generate novel views of dynamic scenes. Access to multiple input views greatly simplifies the task, allowing the scene geometry to be recovered via multi-view stereo~\cite{seitz2006comparison}.

\myparagraph{Novel view synthesis from a single image}
In early works, Debevec \etal \cite{debevec1996modeling} recover 3D scene models and Horry \etal~\cite{horry1997tour} fit a regular mesh to generate novel views. Liebowitz \etal~\cite{liebowitz1999creating} and Criminsi \etal~\cite{criminisi2000single} generate meshes via projective geometry constraints but these methods came at the expense of manual editing.
Hoiem \etal \cite{hoiem2005automatic} generate automatic 3D pop-up by fitting vertical and ground planes onto the 2D image, unlike our mesh representation.
More recently in \cite{niklaus20193d,kopf2020one,shih20203d}, layered depth images are used for single image view synthesis based on a pre-trained depth estimator. In \cite{tucker2020single}, online videos are used to train a scale-invariant MPI representation for view synthesis. SynSin~\cite{wiles2020synsin} synthesizes novel views from a single image with a feature point cloud.
In contrast, we learn to construct scene meshes instead of point clouds and directly map image texture instead of feature vectors to generate novel views from large viewpoint changes.

\myparagraph{Differentiable mesh rendering} Recent work on differentiable mesh renders \cite{kato2018neural,liu2019soft,ravi2020accelerating} allow learning 3D structures through synthesis. NMR \cite{kato2018neural} and SoftRas \cite{liu2019soft} reconstruct the 3D object shape as a mesh by rendering it, comparing it with the input image, and back-propagating losses to refine the mesh geometry. CMR \cite{kanazawa2018learning}, CSM \cite{kulkarni2019canonical} and U-CMR \cite{goel2020shape} build category-specific object meshes from images by deforming from a mean or template category shape through silhouette (and keypoints in \cite{kanazawa2018learning}) supervision. 
Our method is aligned with the analysis-by-synthesis paradigm above. However, unlike most previous works that apply differentiable mesh rendering to objects, we learn the 3D geometry of \textit{scenes} through the rendering losses on the novel view. Moreover, instead of predicting a texture flow as in \cite{kanazawa2018learning,goel2020shape}, we propose to analytically sample the mesh texture from the input view with a differentiable texture sampler. Unlike \cite{pavlakos2019texturepose}, our differentiable texture sampler considers multiple mesh faces in the z-buffer (soft rasterization instead of only the closest one), and assumes perspective (instead of orthographic) camera projection.

\section{\methodname: Rendering the World in a Sheet}
In this work, we propose \textit{\methodname} to synthesize novel views from a single image, as shown in Figure~\ref{fig:method}. Our model build a 3D scene mesh $M$ by warping a lattice grid (\ie a ``sheet'') onto the scene geometry, and is trained with only 2D rendering losses  \textit{without any 3D or depth supervision}.

\subsection{Scene mesh prediction by warping a sheet}
\label{sec:method_mesh_pred}
From the input view image $I_{in}$ of size $W_{im} \times H_{im}$, we build a scene mesh by warping a $W_{m} \times H_{m}$ lattice grid (\ie a sheet) onto the scene, as shown in Figure~\ref{fig:method}. We first extract a $W_{m} \times H_{m}$ visual feature map $\{q_{w,h}\}$ from $I_{in}$ with a convolutional neural network.
Each $q_{w,h}$ is a feature vector at spatial location $(w,h)$ on the $W_{m} \times H_{m}$ network output. In our implementation, we use ResNet-50 \cite{he2016deep} (pretrained on ImageNet) with dilation \cite{yu2015multi} to output features $\{q_{w,h}\}$.

From each $q_{w,h}$ on the feature map, we predict the grid offset $\Delta \hat{x}_{w,h}$ and $\Delta \hat{y}_{w,h}$ to decide how much the vertex $(w,h)$ on the grid should move away from its anchor positions within the image plane (we output $\Delta \hat{x}_{w,h}$ and $\Delta \hat{y}_{w,h}$ in 
NDC space \cite{shreiner2009opengl} between $-1$ to $1$). We also predict how far each vertex is from the camera, \ie its depth $z_{w,h}$. These values are predicted using learned mappings as
\begin{eqnarray}
\Delta \hat{x}_{w,h} &=& \tanh\left(W_1 q_{w,h} + b_1 \right) / (W_{m} - 1) \label{eqn:offset_x}\\
\Delta \hat{y}_{w,h} &=& \tanh\left(W_2 q_{w,h} + b_2 \right) / (H_{m} - 1) \label{eqn:offset_y} \\
z_{w,h} &=& g\left(W_3 q_{w,h} + b_3\right) \label{eqn:depth}
\end{eqnarray}
where division by $(W_{m} - 1)$ and $(H_{m} - 1)$ ensures that the vertices can only move within a certain range. $g(\cdot)$ is a scalar nonlinear function to scale the network prediction into depth values. We use $g(\psi) = \alpha_g / (\sigma(\psi) + \epsilon_g) + \beta_g$ in our implementation, where $\sigma(\cdot)$ is the sigmoid function and $\alpha_g$, $\beta_g$ and $\epsilon_g$ are fixed hyper-parameters.

\myparagraph{Building the 3D scene mesh} We first build the mesh vertices $\{V_{w,h}\}$ from the grid offset and depth as
\begin{equation}
\label{eqn:mesh}
    V_{w,h} = 
                        \begin{bmatrix}
    z_{w,h} \cdot (\hat{x}_{w,h} + \Delta \hat{x}_{w,h}) \cdot \tan(\theta_F / 2) \\
    z_{w,h} \cdot (\hat{y}_{w,h} + \Delta \hat{y}_{w,h}) \cdot \tan(\theta_F / 2) \\
    z_{w,h}\\
    \end{bmatrix}
\end{equation}
for $w=1,\cdots,W_{m}$ and $h=1,\cdots,H_{m}$. Here $\theta_F$ is the camera field-of-view, and $\hat{x}_{w,h}$ and $\hat{y}_{w,h}$ are anchor positions on the grid equally spaced from $-1$ to $1$.

Then, we connect the mesh vertices $\{V_{w,h}\}$ along the edges on the grid to form mesh faces $\{F\}$ as shown in Figure~\ref{fig:method} and obtain a 3D mesh $M = (\{V_{w,h}\}, \{F\})$. A vertex in the mesh is connected to its 4 or 8 neighbours on the grid. 

To encourage the mesh surface to be smooth unless it needs to bend to fit the scene geometry, we apply a Laplacian term $L_{m} = \sum_{w,h} \left\| \sum_{(\bar{w},\bar{h}) \in N(w,h)} \left( V_{\bar{w},\bar{h}} - V_{w,h} \right) \right\|_1$ on the mesh vertices, where $N(w,h)$ are the adjacent vertices to $(w, h)$. In addition, we also apply an L2 regularization term $L_{g} = \sum_{w,h} \left(\Delta \hat{x}_{w,h}^2 + \Delta \hat{y}_{w,h}^2\right)$ to the grid offset.

\subsection{Differentiable texture sampler}  \label{sec:method_texture_sampling}

To render the input scene in another camera pose for novel view synthesis, we need to project image texture from the input view to the target view in a differentiable manner. While existing renderers \cite{kato2018neural,liu2019soft,ravi2020accelerating} can render an image from a scene mesh based on its texture map, they cannot directly transform image pixels in screen space \textit{between} two different camera poses. In our model, we accomplish differentiable projection between two views by \textit{first reconstructing the scene mesh's texture map from the input view} (which involves inverting the texture-map-to-image perspective transform in a differentiable manner) so that it can be later rendered with the scene mesh in novel views using existing mesh renderers.

While a few approaches \cite{kanazawa2018learning,goel2020shape} build a mesh texture map with a learned texture flow on objects, it is hard to apply the same to scenes, which do not have canonical shapes. Here, we take an alternative route and propose a differentiable texture sampler. 
We \textit{analytically} sample the mesh texture $\hat{T}$ as a UV texture map \cite{shreiner2009opengl} from the input view $I_{in}$, where gradients $\partial \hat{T} / \partial V$ and $\partial \hat{T} / \partial I_{in}$ over the vertex coordinates and the input image respectively can be computed.

To implement this texture sampler, we project the mesh faces onto the image plane to build a buffer (sorted in ascending z-order) containing the z values and 2D euclidean distance of points on the closest $K$ mesh faces whose projection overlaps image pixel $p_{i, j}$ as in PyTorch3D \cite{ravi2020accelerating}. Then, we splat the RGB pixel intensities from the image $I_{in}$ onto the UV texture map $\hat{T}$. Specifically, we first compute the weight $w_{i, j}^{k}$ denoting the contribution of the $k$-th face color on pixel $p_{i, j}$ based on the softmax blending formulation in \cite{ravi2020accelerating, liu2019soft}. 
We then decompose the input image $I_{in}$ into $K$ images $I^{k}_{in}$, where $I^{k}_{in}(i, j) = I_{in} \cdot w_{i, j}^{k}$, and splat the RGB pixels from each $I^{k}_{in}$ to a texture map layer $\hat{T}^k$ as
\begin{equation}
    \hat{T}^k = \mathrm{splat}(I^{k}_{in}, f^k)
\end{equation}
where the flow $(u,v)=f^k(i, j)$ maps image coordinates $(i, j)$ to UV coordinates $(u,v)$ on the $k$-th mesh face in the z-buffer. Here $\mathrm{splat}$ is a differentiable splatting operation from the image space $I^{k}_{in}$ to the texture space $\hat{T}^k$. Finally, we sum all the $K$ texture maps as the final UV texture map $\hat{T} = \sum_{k}\hat{T}^k$. Please see supplemental for more details.

In summary, the image pixels are splatted onto the texture space via each rasterized mesh face, and blended together to obtain the final texture map. The entire process is differentiable with respect to both $I_{in}$ and the mesh vertex coordinates $\{V\}$, as one can analytically compute $\partial \hat{T}^k/\partial I^k_{in}$, $\partial \hat{T}^k/\partial f^k$, $\partial f^k/\partial V$, and $\partial w_{i, j}^{k}/\partial V$.

\subsection{Learning scene geometry by view synthesis}
\label{sec:method_analysis_by_synthesis}

\begin{figure}[t]
\centering
\includegraphics[width=\linewidth]{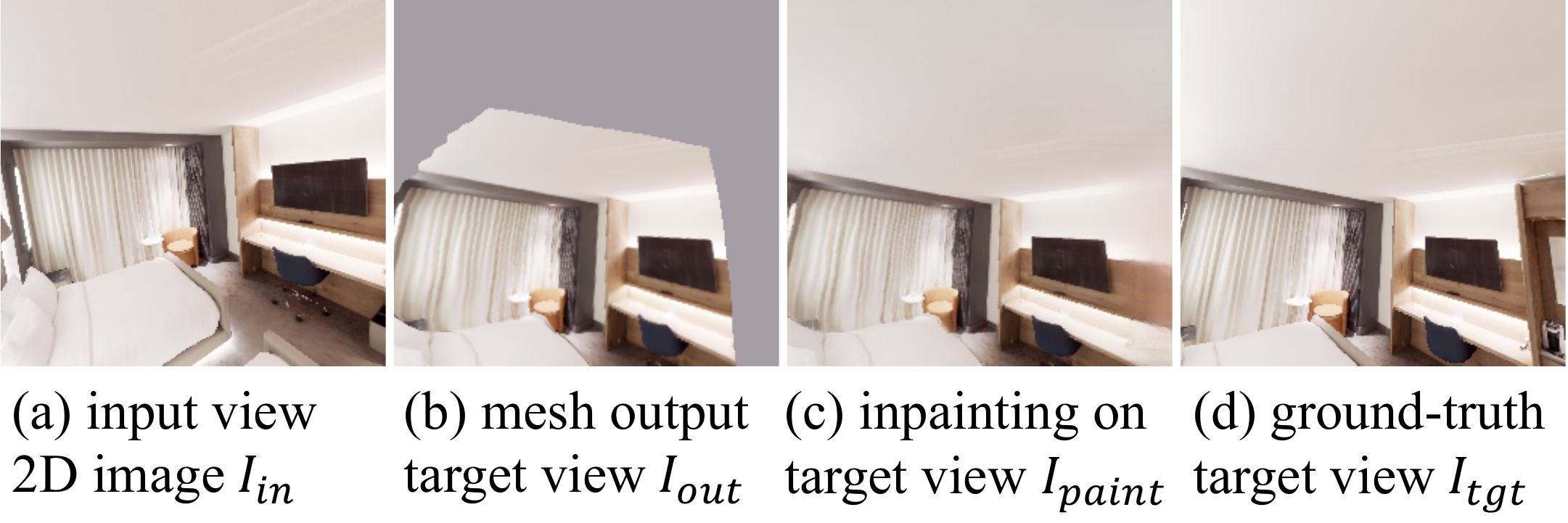}
\vspace{-2em}
\caption{Parts of the target view (the grey area in b) are often invisible from the input and must be imagined based on prior knowledge. We make plausible predictions over the invisible regions with an inpainting network (c). However, this task is inherently uncertain (\eg one cannot be sure about the rightmost cabinet in d).}
\label{fig:seen_vs_hallucinated}
\vspace{-1.3em}
\end{figure}

To synthesize a novel view, we project the mesh vertex coordinates $\{V\}$ from the input camera pose $\theta_{in}$ to $\{V^{tgt}\}$ in the camera coordinate space of the target viewpoint $\theta_{tgt}$. Then, we render the mesh $M^{tgt} = (\{V^{tgt}\}, \{F\})$ in the target camera pose along with its texture map $\hat{T}$ to output a 2D image $I_{out}$ of size $W_{im} \times H_{im}$ as the target view:
\begin{equation}
\label{eqn:render_mesh}
    I_{out} = \mathrm{render\_mesh}(\{V^{tgt}\}, \{F\}, \hat{T}).
\end{equation}
We use the differentiable mesh renderer in \cite{ravi2020accelerating} so that we can compute the gradients $\partial I_{out} / \partial V^{tgt}$ and $\partial I_{out} / \partial \hat{T}$. Through mesh rendering, we also obtain a foreground mask $F_{out}$ with the same size as $I_{out}$, indicating which pixels in the rendered image $I_{out}$ are covered by the mesh and which pixels are from background color, as shown by the grey area in Figure~\ref{fig:seen_vs_hallucinated} (b).

Our model is supervised with paired input and target views of a scene (along with their camera poses). We use a pixel L1 loss $L^{rgb}_{out} = \|I_{out} - I_{tgt}\|_1 / (W_{im} \cdot H_{im}) \label{eqn:loss_rgb}$ and a perceptual loss~\cite{wang2018high,wiles2020synsin} $L^{pc}_{out} = P(I_{out}, I_{tgt})$, where $I_{tgt}$ is the ground-truth target view image. The model then needs just a single image at test time.

\subsection{Inpainting and image refinement}
\label{sec:method_inpainting}

\begin{table*}[t]
\vspace{-1em}
\footnotesize
\begin{center}
\setcounter{magicrownumbers}{0}
\begin{tabular}{clcccccccccccc}
\toprule
& &\multicolumn{9}{c}{Matterport \cite{chang2017matterport3d}} & \multicolumn{3}{c}{Replica \cite{straub2019replica}} \\
 \cmidrule(lr){3-11} \cmidrule(lr){12-14}
& & \multicolumn{3}{c}{PSNR $\uparrow$} & \multicolumn{3}{c}{SSIM $\uparrow$} & \multicolumn{3}{c}{Perc Sim $\downarrow$} & PSNR $\uparrow$ & SSIM $\uparrow$ & Perc Sim $\downarrow$ \\ 
\# & Method & Both & InVis & Vis & Both & InVis & Vis & Both & InVis & Vis & \\ 
\cmidrule(lr){1-2} \cmidrule(lr){3-5} \cmidrule(lr){6-8} \cmidrule(lr){9-11} \cmidrule(lr){12-14}
\rownumber & Im2Im \cite{zhou2016view} & 15.87 & 16.20 & 15.97 & 0.53 & 0.60 & 0.48 & 2.99 & 0.58 & 2.05 & 17.42 & 0.66 & 2.29 \\
\rownumber & Tatarchenko \etal \cite{tatarchenko2016multi} & 14.79 & 14.83 & 15.05 & 0.57 & 0.62 & 0.53 & 3.73 & 0.74 & 2.50 & 14.36 & 0.68 & 3.36 \\
\rownumber & Vox \cite{sitzmann2019deepvoxels} w/ UNet & 18.52  & 17.85 & 19.05 & 0.57 & 0.57 & 0.57 & 2.98 & 0.77  & 1.96 & 18.69 & 0.71 & 2.68 \\
\rownumber & Vox \cite{sitzmann2019deepvoxels} w/ ResNet & 20.62 & 19.64 & 21.22 & 0.70 & 0.69 & 0.68 & 1.97 & 0.47 & 1.19 & 19.77 & 0.75 & 2.24 \\
\rownumber & SynSin \cite{wiles2020synsin} & 20.91 & 19.80 & 21.62 & 0.71 & 0.71 & 0.70 & 1.68 & 0.43 & 0.99 & 21.94 & 0.81 & 1.55 \\
\midrule
\rownumber & ours w/o inpainting & -- & -- & 25.42 & -- & -- & 0.80 & -- & -- & 0.68 & -- & -- & -- \\
\rownumber & ours & \textbf{24.67} & \textbf{22.90} & \textbf{26.00} & \textbf{0.82} & \textbf{0.77} & \textbf{0.82} & \textbf{1.05} & \textbf{0.35} & \textbf{0.54} & \textbf{23.51} & \textbf{0.85} & \textbf{1.32} \\
\bottomrule
\end{tabular}
\end{center}
\vspace{-2em}
\caption{{Novel View Synthesis}: Performance of our and previous approaches on the Matterport dataset and the Replica dataset. All models are trained on Matterport and evaluated on both datasets. See Sec.~\ref{sec:exp_mp3d_replica} for details.}
\label{tab:exp_mp3d_replica}
\vspace{-1.5em}
\end{table*}

The target view image consists of two parts: things that can be directly seen from the input view $I_{in}$, and things that need to be imagined based on our prior knowledge of the visual world, as illustrated in Figure~\ref{fig:seen_vs_hallucinated}. As our mesh warping and rendering procedure in Sec.~\ref{sec:method_mesh_pred}, \ref{sec:method_texture_sampling} and \ref{sec:method_analysis_by_synthesis} builds a pixel-to-pixel correspondence between the input and the target view, it only renders pixels that are \textit{visible} from the input view. To obtain a plausible imagination of the \textit{invisible} image regions, we apply an inpainting network $G$ on the rendered mesh $I_{out}$ to fill the missing regions and output a new image $I_{paint} = G(I_{out})$ as the final target view.

We build our inpainting network based on the generator in pix2pixHD \cite{wang2018high}, which translates a 4-channel input (the rendered image $I_{out}$ and its foreground mask $F_{out}$) into a 3-channel output image $I_{paint}$. Our inpainting network outputs an entire image -- it not only fills the invisible regions but also refines the image details in the visible regions. We apply the same RGB pixel L1 loss $L^{rgb}_{paint}$ and perceptual loss $L^{pc}_{paint}$ as in Sec.~\ref{sec:method_analysis_by_synthesis} on the inpainting output $I_{paint}$.

\myparagraph{Training} We train our model using the Adam optimizer \cite{kingma2014adam} with a weighted combination of losses as
    $L = \lambda_1 L^{rgb}_{out} + \lambda_2 L^{pc}_{out} + \lambda_3 L^{rgb}_{paint} + \lambda_4 L^{pc}_{paint} + \lambda_5 L_{g} + \lambda_6 L_{m} \nonumber$
with $\lambda_1 = \lambda_3 = 8$, $\lambda_2 = \lambda_4 = 2$, $\lambda_5 = 0.2$, and $\lambda_6 = 10^{-4}$. Our model is trained for a total of 50000 iterations with batch size $64$ and $10^{-4}$ learning rate.

We use a grid mesh with size $W_{m} \times H_{m} = 33 \times 33$ (and also $65 \times 65$ in Sec.~\ref{sec:exp_realestate10k}). Following SynSin \cite{wiles2020synsin}, we use $W_{im} \times H_{im} = 256 \times 256$ as the input and output image size. Our mesh implementation is based on PyTorch3D \cite{ravi2020accelerating}.

\subsection{Extension: multi-layered Worldsheets}
\label{sec:method_multiple_layers}
Although shrink-wrapping a single mesh sheet onto images works well on a wide range of scenes, one limitation is that it assumes that the foreground objects are connected to the background by mesh faces, which sometimes causes artifacts near object boundaries or depth discontinuities.

We propose an extension to address this limitation: predicting and warping multiple layers of Worldsheet onto the scene, where each sheet has a transparency channel in its texture map, loosely inspired by layered-depth images~\cite{shade1998layered}. This allows some layers to fit the foreground object and others to capture the background. Specifically, we predict grid offset and depth for each mesh sheet from the feature map $\{q_{w,h}\}$ following Eqn.~\ref{eqn:offset_x} to \ref{eqn:depth} with separate parameters. We also predict an $H_{im} \times W_{im}$ alpha map for each sheet using a deconvolution layer on $\{q_{w,h}\}$, which is then projected to the transparency channel in the UV texture map of the associated sheet. Finally, the multiple mesh sheets are rendered in the novel view using alpha compositing \cite{smith1995image}. The whole model can be trained end-to-end under the same supervision. In Sec.~\ref{sec:exp_multiple_layers}, we find that, qualitatively, this extension leads to better handling of occlusions and parallax effect than a single mesh sheet.

\section{Experiments}

\begin{figure*}[t]
\vspace{-1.5em}
\small
\centering
\begin{tabular}{ccc}
\includegraphics[height=210pt,trim=0 0 825pt 0, clip]{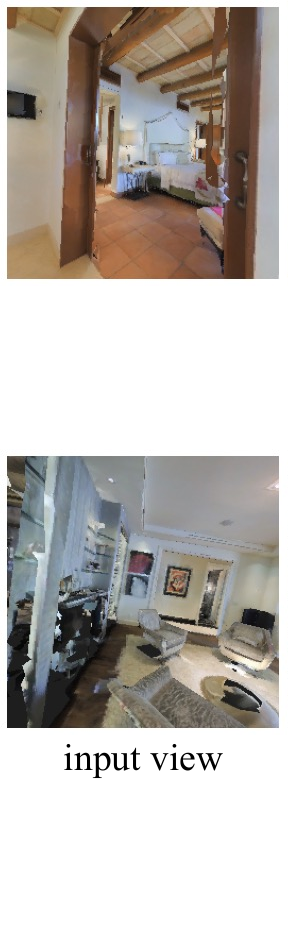} &
\includegraphics[height=210pt,trim=138pt 0 276pt 0, clip]{mp3d_vis.pdf} &
\includegraphics[height=210pt,trim=690pt 0 0 0, clip]{mp3d_vis.pdf} \\
\end{tabular}
\vspace{-5.0em}
\caption{Novel views from our and previous methods on the Matterport dataset (scene mesh shown in the last column). The first row have same viewpoint change as in \cite{wiles2020synsin} while the second row has $2\times$ larger camera angle change.}
\label{fig:vis_mp3d}
\vspace{-0.5em}
\end{figure*}

\begin{table*}[t]
\footnotesize
\begin{center}
\setcounter{magicrownumbers}{0}
\begin{tabular}{clcccccccccccc}
\toprule
& &\multicolumn{9}{c}{Matterport \cite{chang2017matterport3d} ($2\times$ cam. change)} & \multicolumn{3}{c}{Replica \cite{straub2019replica} ($2\times$ cam. change)} \\
 \cmidrule(lr){3-11} \cmidrule(lr){12-14}
& & \multicolumn{3}{c}{PSNR $\uparrow$} & \multicolumn{3}{c}{SSIM $\uparrow$} & \multicolumn{3}{c}{Perc Sim $\downarrow$} & PSNR $\uparrow$ & SSIM $\uparrow$ & Perc Sim $\downarrow$ \\ 
\# & Method & Both & InVis & Vis & Both & InVis & Vis & Both & InVis & Vis & \\ 
\cmidrule(lr){1-2} \cmidrule(lr){3-5} \cmidrule(lr){6-8} \cmidrule(lr){9-11} \cmidrule(lr){12-14}
\rownumber & Im2Im \cite{zhou2016view} & 14.93 & 15.16 & 15.28 & 0.51 & 0.56 & 0.46 & 3.26 & 0.93 & 1.91 & 15.91 & 0.63 & 2.63 \\
\rownumber & Tatarchenko \etal \cite{tatarchenko2016multi} & 14.71 & 14.77 & 15.08 & 0.56 & 0.61 & 0.52 & 3.74 & 1.04 & 2.14 & 14.19 & 0.68 & 3.37 \\
\rownumber & SynSin \cite{wiles2020synsin} & 19.15 & 17.76 & 20.69 & 0.67 & 0.66 & 0.66 & 2.06 & 0.78 & 0.96 & 19.63 & 0.77 & 1.94 \\
\midrule
\rownumber & ours w/o inpainting & -- & -- & 24.20 & -- & -- & 0.76 & -- & -- & 0.69 & -- & -- & -- \\
\rownumber & ours & \textbf{22.62} & \textbf{20.89} & \textbf{24.76} & \textbf{0.77} & \textbf{0.72} & \textbf{0.77} & \textbf{1.41} & \textbf{0.63} & \textbf{0.56} & \textbf{21.12} & \textbf{0.81} & \textbf{1.70} \\
\bottomrule
\end{tabular}
\end{center}
\vspace{-2em}
\caption{{Novel View Synthesis}: Generalization performance to larger viewpoint changes of our model vs. previous approaches on the Matterport dataset and the Replica dataset. All models are trained on the Matterport dataset and evaluated on both datasets with $2\times$ larger camera angle changes than in the training data. See Sec.~\ref{sec:exp_mp3d_replica} for details.}
\label{tab:exp_mp3d_replica_2x_cam}
\vspace{-1.5em}
\end{table*}

We evaluate our model on three datasets: Matterport \cite{chang2017matterport3d}, Replica~\cite{straub2019replica}, and RealEstate10K \cite{zhou2018stereo}, following the experimental setup and details from ~\cite{wiles2020synsin}.
We then provide analysis on in-the-wild images and multi-layered sheets.

\subsection{Evaluation on Matterport and Replica}
\label{sec:exp_mp3d_replica}
We first train and evaluate our approach on the Matterport dataset \cite{chang2017matterport3d}, which contains 3D scans of homes. We load the Matterport dataset in the Habitat simulator \cite{savva2019habitat}, following the same training, validation, and test splits as in SynSin~\cite{wiles2020synsin}. During training, we supervise our model with paired 2D images of the input and the target views. We empirically find that it works slightly better to first train the scene mesh predictor (Sec.~\ref{sec:method_mesh_pred}) and then freeze the scene mesh to further train the inpainting network (Sec.~\ref{sec:method_inpainting}), rather than training both components jointly from scratch.

\myparagraph{Metrics} Following SynSin \cite{wiles2020synsin}, we evaluate the predicted novel view images $I_{paint}$ using three metrics: Peak Signal-to-Noise Ratio (\textbf{PSNR}; higher is better), Structural Similarity (\textbf{SSIM}; higher is better), and Perceptual Similarity distance (\textbf{Perc Sim}; lower is better). The Perc Sim metric is based on the convolutional feature distance between the prediction and the ground-truth, which is shown to be highly correlated with human judgement \cite{zhang2018unreasonable,wiles2020synsin}. Since only a part of the target view image can be seen from the input image as illustrated in Figure~\ref{fig:seen_vs_hallucinated}, we separately evaluate these metrics on visible regions (\textbf{Vis}, which can be seen from the input view), invisible regions (\textbf{InVis}, which cannot be seen and must be imagined), and the entire image (\textbf{Both}). Note that the visible region masks are obtained from the ground-truth scene geometry and camera frustum (available from the Habitat simulator) instead of predicted by our mesh, and are the same as in SynSin's evaluation.

\myparagraph{Baselines} We compare our method to several previous approaches: \textbf{Im2Im} \cite{zhou2016view} is an image-to-image translation method which predicts an appearance flow to warp an input view to the target view based on an input camera transformation. \textbf{Tatarchenko \etal} \cite{tatarchenko2016multi} is similar to Im2Im, but directly predicts the target view image instead of an appearance flow. \textbf{Vox w/ UNet} and \textbf{Vox w/ ResNet} are two variants of the deep voxel representation \cite{sitzmann2019deepvoxels} with different encoder-decoder architectures based on UNet, or ResNet as implemented in \cite{wiles2020synsin}. \textbf{SynSin} \cite{wiles2020synsin} projects a dense feature point cloud (extracted from every image pixel) to the target camera pose and applies a refinement network on the point cloud projection to output the target view image. 
We also evaluate the prediction of our model before inpainting (\ie directly using the mesh rendering output $I_{out}$ as the target view) to analyze how well our method performs with texture sampling and mesh rendering alone.

\myparagraph{Results} The results are shown in Table~\ref{tab:exp_mp3d_replica}. Even without inpainting, the mesh rendering output $I_{out}$ from our method already outperforms previous approaches by a large margin under all the three metrics on the visible regions. With the help of an inpainting network, our final output $I_{paint}$ has significantly higher performance than previous work on both invisible and visible regions, achieving a new state-of-the-art performance on this dataset. Figure~\ref{fig:vis_mp3d} shows view synthesis examples from our method and previous work on the Matterport dataset, where our method can paint things such as doorframe or sofa at more precise locations.

\begin{figure}[t]
\footnotesize
\centering
\includegraphics[width=\linewidth]{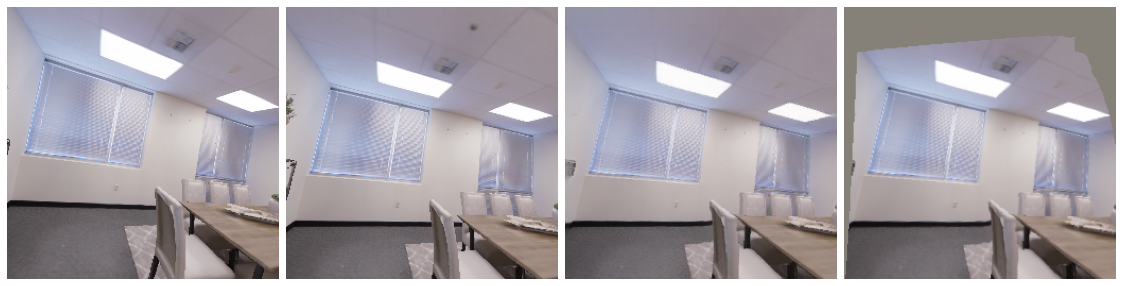} \\
\begin{tabular}{cccc}
~~~~input view & ~~~GT target view & ~~prediction & pred. w/o inpainting \\
\end{tabular}
\vspace{-2em}
\caption{Generalization of our model (trained on Matterport) to the Replica dataset without retraining (Sec.~\ref{sec:exp_mp3d_replica}).}
\label{fig:vis_replica}
\vspace{-1.5em}
\end{figure}

\begin{figure*}[t]
\vspace{-1.5em}
\small
\centering
\begin{tabular}{ccc}
\includegraphics[height=210pt,trim=0 0 825pt 0, clip]{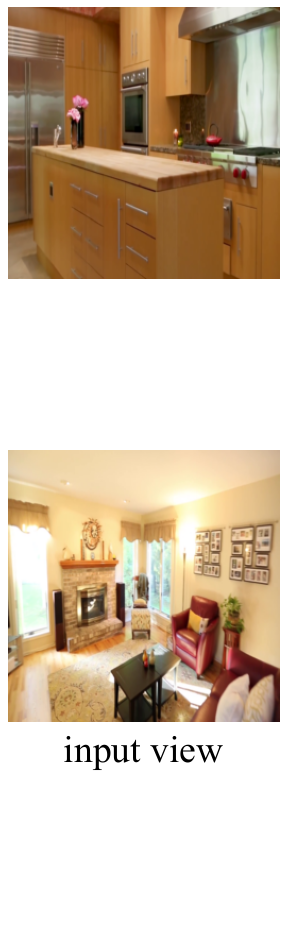} &
\includegraphics[height=210pt,trim=138pt 0 276pt 0, clip]{realestate10k_vis.pdf} &
\includegraphics[height=210pt,trim=690pt 0 0 0, clip]{realestate10k_vis.pdf} \\
\end{tabular}
\vspace{-5.0em}
\caption{Novel views from our and previous methods on the RealEstate10K dataset (scene mesh shown in the last column).}
\label{fig:vis_realestate10k}
\vspace{-0.5em}
\end{figure*}

\myparagraph{Generalization to the Replica dataset} Following \cite{wiles2020synsin}, we also evaluate how well our model generalizes to another scene dataset, Replica \cite{straub2019replica}, which contains high-quality laser scans of both homes and offices. We take our model trained on the Matterport dataset and directly evaluate on the Replica dataset without re-training. The results are shown in Table~\ref{tab:exp_mp3d_replica}, where all methods are trained and evaluated under the same setting. It can be seen that our method achieves noticeably better generalization to this dataset and outperforms previous approaches by a large margin. Figure~\ref{fig:vis_replica} shows view synthesis examples on the Replica dataset.

\myparagraph{Generalization to larger viewpoint changes} We further analyze how well our and previous approaches generalize to larger camera pose changes beyond their training data. In this analysis, we sample new input-target view pairs on the test scenes with $2\times$ larger camera angle changes than in the training data, and directly evaluate all approaches on these new viewpoints without retraining. The results are shown in Table~\ref{tab:exp_mp3d_replica_2x_cam}, where our method largely outperforms other approaches under all metrics. Figure~\ref{fig:vis_mp3d} (second row) shows an example under $2\times$ larger camera angle change.

\subsection{Evaluation on RealEstate10K}
\label{sec:exp_realestate10k}

The RealEstate10K dataset \cite{zhou2018stereo} consists of both indoor and outdoor scenes extracted from YouTube videos of houses. The input view and the target view are different video frames within a time range, with camera poses estimated using structure-from-motion.

\begin{table}[t]
\vspace{-1em}
\footnotesize
\begin{center}
\setcounter{magicrownumbers}{0}
\begin{tabular}{clccc}
\toprule
\# & Method & PSNR $\uparrow$ & SSIM $\uparrow$ & Perc Sim $\downarrow$ \\ 
\cmidrule(lr){1-2} \cmidrule(lr){3-3} \cmidrule(lr){4-4} \cmidrule(lr){5-5}
\rownumber & Im2Im \cite{zhou2016view} & 17.05 & 0.56 & 2.19 \\
\rownumber & Tatarchenko \etal \cite{tatarchenko2016multi} & 11.35 & 0.33 & 3.95 \\
\rownumber & Vox \cite{sitzmann2019deepvoxels} w/ UNet & 17.31 & 0.53 & 2.30 \\
\rownumber & Vox \cite{sitzmann2019deepvoxels} w/ ResNet & 21.88 & 0.71 & 1.30 \\
\rownumber & 3DView (similar to \cite{3dview}) & 21.88 & 0.66 & 1.52 \\
\rownumber & SynSin~\cite{wiles2020synsin} & 22.83 & 0.75 & 1.13 \\
\rownumber & Single-View MPI~\cite{tucker2020single} & 24.03 & 0.78 & 1.18 \\
\rownumber & StereoMag~\cite{zhou2018stereo} & 25.34 & \textbf{0.82} & 1.19 \\
\midrule
\rownumber & ours ($33\times 33$ mesh) & 26.24 & \textbf{0.82} & 0.83 \\
\rownumber & ours ($65\times 65$ mesh) & \textbf{26.74} & \textbf{0.82} & \textbf{0.80} \\
\bottomrule
\end{tabular}
\end{center}
\vspace{-2em}
\caption{Comparison of our model with previous work on the RealEstate10K dataset \cite{zhou2018stereo}. See Sec.~\ref{sec:exp_realestate10k} for details.}
\label{tab:realestate10k}
\vspace{-1.5em}
\end{table}

On this dataset, we follow the experimental setup in SynSin \cite{wiles2020synsin} and use the same training, validation, and test data. In addition to using a $33 \times 33$ mesh, we also train our model with a higher resolution $W_{m} \times H_{m} = 65 \times 65$ mesh, which is initialized from a trained $33 \times 33$ mesh model with a new transposed convolution layer to upsample the feature map $\{q_{w,h}\}$ in Sec.~\ref{sec:method_mesh_pred} to $65 \times 65$ spatial dimensions.

We compare our method to several previous approaches. In addition to the baselines in Sec.~\ref{sec:exp_mp3d_replica}, we also compared to three additional approaches. \textbf{3DView} is a system similar to the Facebook 3D Photo \cite{3dview} based on layered depth images and is also a baseline in \cite{wiles2020synsin}. \textbf{Single-View MPI} \cite{tucker2020single} and \textbf{StereoMag} \cite{zhou2018stereo} both use multiplane images (MPIs), where Single-View MPI builds MPIs from a single input image while StereoMag relies on a stereo pair using images from two different views as input at test time. Except for StereoMag, all other methods use a single view at test time.

\begin{figure*}[t]
\vspace{-1em}
\small
\centering
\includegraphics[width=0.31\textwidth]{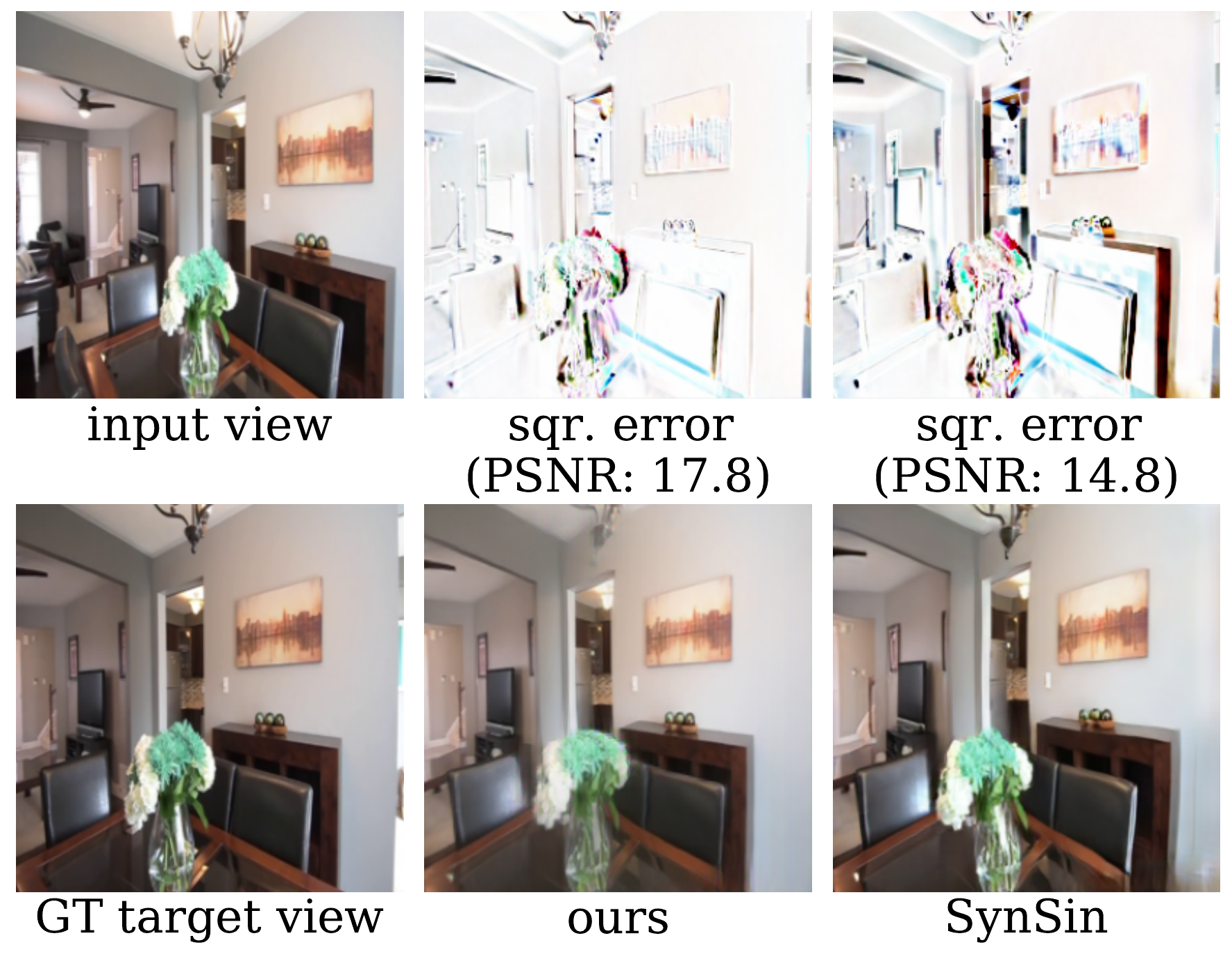}~~~~~~~~
\includegraphics[width=0.31\textwidth]{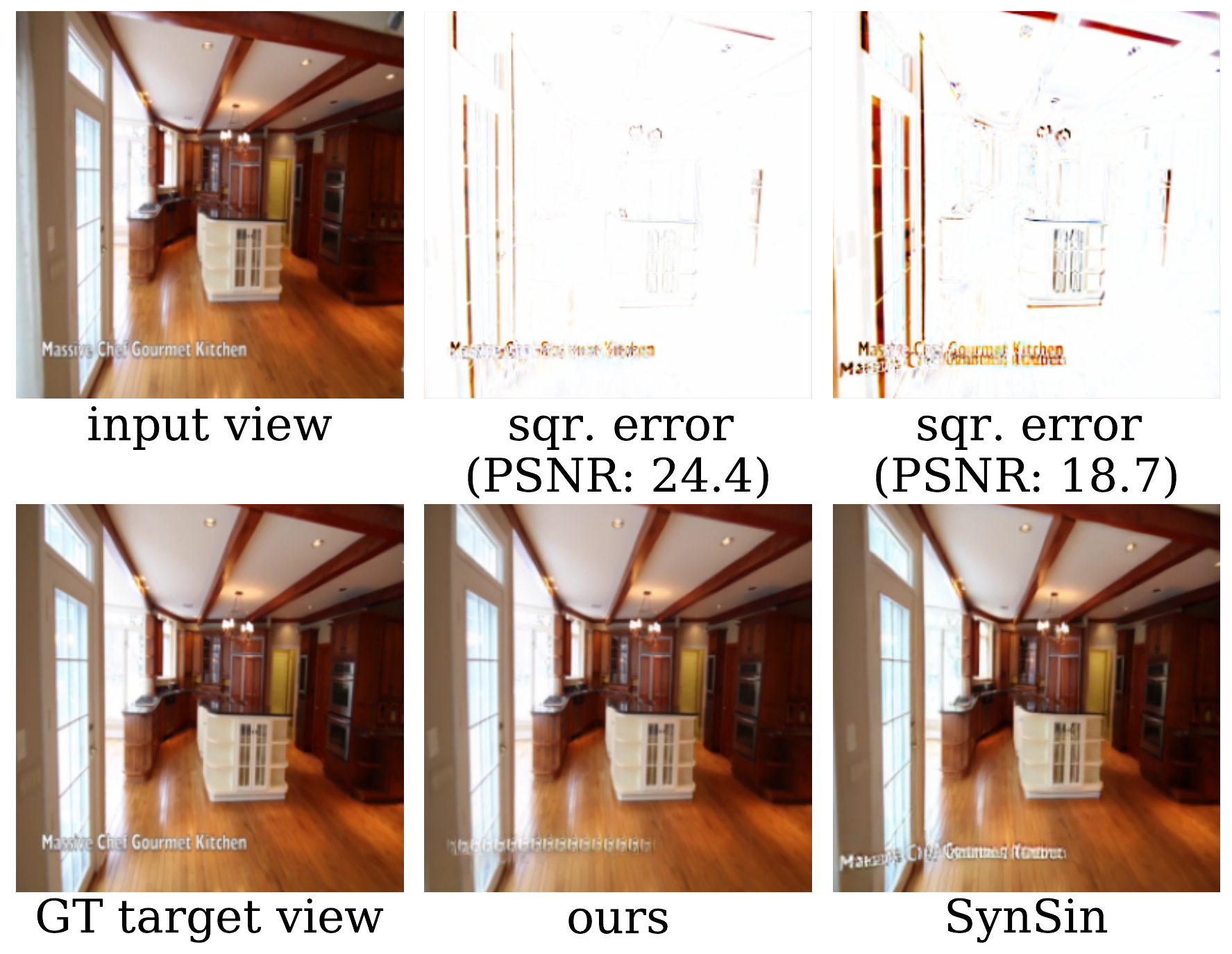}~~~~~~~~
\includegraphics[width=0.31\textwidth]{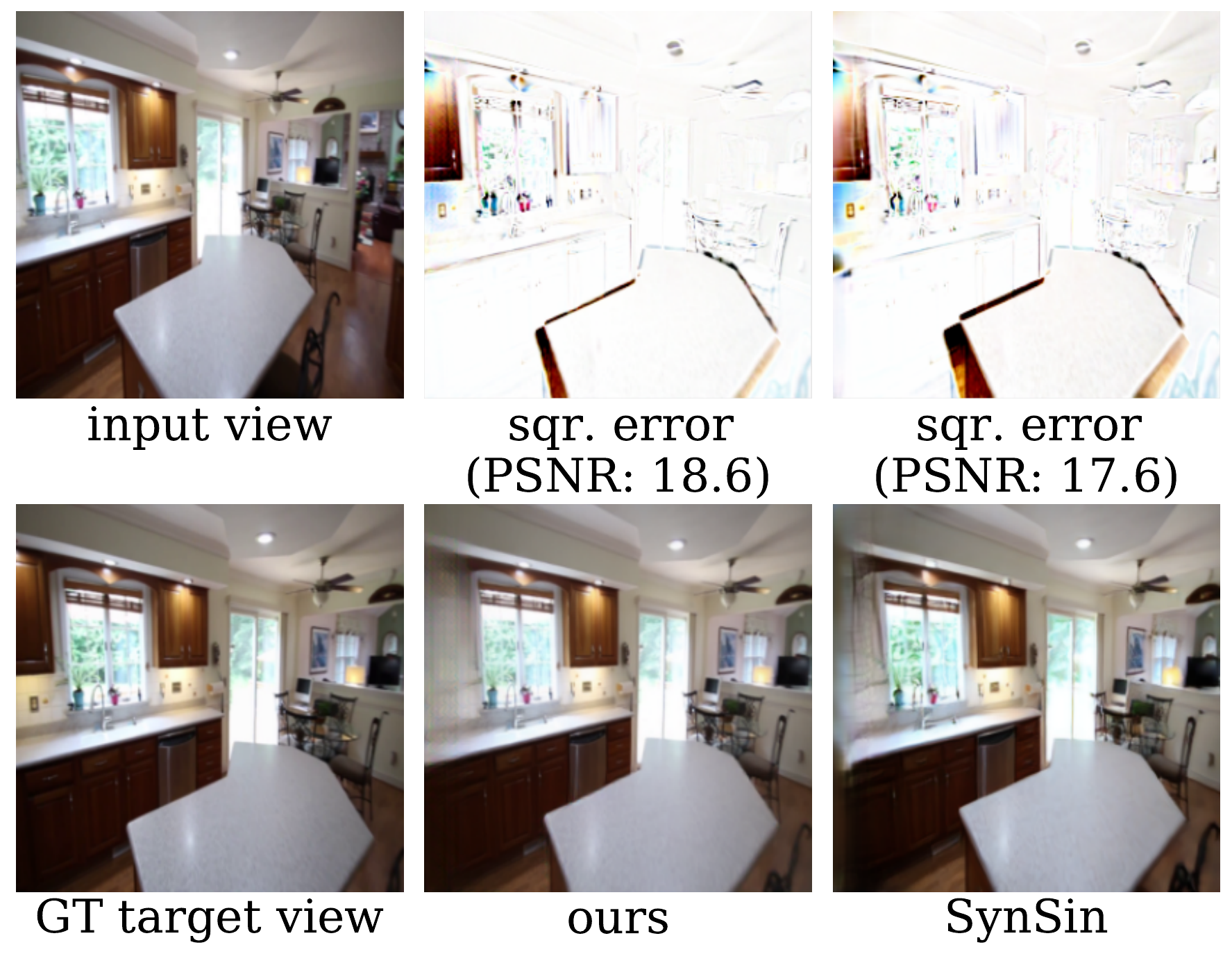}
\vspace{-2em}
\caption{Squared error maps in target view from our method and SynSin \cite{wiles2020synsin} on the RealEstate10K dataset (darker is higher error). Our method paints things in the novel view at more precise locations, resulting in lower error and higher PSNR.}
\label{fig:vis_realestate10k_psnr}
\vspace{-1.5em}
\end{figure*}

\begin{table}[t]
\vspace{-1em}
\footnotesize
\begin{center}
\setcounter{magicrownumbers}{0}
\begin{tabular}{clccccccccc}
\toprule
\# & Ablation setting & PSNR $\uparrow$ & SSIM $\uparrow$ & Perc Sim $\downarrow$ \\ 
\cmidrule(lr){1-2} \cmidrule(lr){3-3} \cmidrule(lr){4-4} \cmidrule(lr){5-5}
\multirow{2}{*}{\rownumber} & default ($33\times 33$ mesh & \multirow{2}{*}{26.24} & \multirow{2}{*}{\textbf{0.82}} & \multirow{2}{*}{0.83} \\
& ~~reg. weight: $10^{-4}$) \\
\midrule
\rownumber & reg. weight: $0$ & 25.78 & 0.81 & 0.86 \\
\rownumber & reg. weight: $10^{-5}$ & 26.18 & \textbf{0.82} & 0.83\\
\rownumber & reg. weight: $10^{-3}$ & 24.83 & 0.78 & 0.96 \\
\midrule
\rownumber & $5\times 5$ mesh & 24.39 & 0.79 & 0.99 \\
\rownumber & $9\times 9$ mesh & 25.10 & 0.80 & 0.92 \\
\rownumber & $17\times 17$ mesh & 25.91 & 0.81 & 0.84 \\
\rownumber & $65\times 65$ mesh\dag & \textbf{26.74} & \textbf{0.82} & \textbf{0.80} \\
\bottomrule
\end{tabular}
\end{center}
\vspace{-2em}
\caption{Ablations on the RealEstate10K dataset \cite{zhou2018stereo}. See Sec.~\ref{sec:exp_realestate10k} for details. (\dag: initialized from the default model.)}
\label{tab:realestate10k_details}
\vspace{-1.5em}
\end{table}

\myparagraph{Results}
We follow the evaluation protocol of \cite{wiles2020synsin} on RealEstate10K, with results\footnote{To compare with StereoMag \cite{zhou2018stereo} that uses two input views, in the evaluation protocol of SynSin \cite{wiles2020synsin} on RealEstate10K, the best metrics of two separate predictions based on each view were reported for single-view methods. We follow this evaluation protocol for consistency with \cite{wiles2020synsin} on RealEstate10K in Table~\ref{tab:realestate10k} and \ref{tab:realestate10k_details}. We also report averaged metrics over all predictions in supplemental, where the trends are consistent.} shown in Table~\ref{tab:realestate10k}. 
It can be seen that our method achieves the highest performance, outperforming previous approaches by a noticeable margin. Besides, a higher resolution $65 \times 65$ mesh gives a further performance boost. 
Figure~\ref{fig:vis_realestate10k} shows predicted novel views on this dataset. In addition, we visualize the pixel-wise squared error map on the prediction from our method and SynSin \cite{wiles2020synsin} in Figure~\ref{fig:vis_realestate10k_psnr}, where our method paints objects at more precise locations compared to SynSin, resulting in higher PSNR and better quality.

\myparagraph{Ablations} Since our model relies on deforming a mesh sheet, we first analyze the impact of geometric regularization on the mesh deformation. In Table~\ref{tab:realestate10k_details}, line 2 to 4 vary the weight of the mesh Laplacian regularization term $L_m$ from its default value $10^{-4}$. Comparing these variants to line 1, a higher regularization ($10^{-3}$, line 4) restricts the model's capacity to precisely fit the scene geometry and hence hurts the performance. Meanwhile, there is only a smaller drop when decreasing this regularization weight to $10^{-5}$ or even zero, suggesting that our differentiable rendering pipeline provides robust wrapping of the mesh onto the scene.

We further study the impact of mesh resolution $W_{m} \times H_{m}$ in line 5 to 8. As expected, higher mesh resolution allows fitting more fine-grained scene details and gives higher view synthesis performance, with the final $65\times 65$ mesh giving the best performance. In addition, we find that with enough mesh resolution (such as $65\times 65$), one can restrict the grid offset in Eqn.~\ref{eqn:offset_x} and \ref{eqn:offset_y} to zero and only use the predicted depth in Eqn.~\ref{eqn:depth} to deform the mesh, which gives only $-0.13$ PSNR drop (however, the grid offset makes a larger difference in lower-resolution meshes such as Figure~\ref{fig:method}).

\subsection{Analysis: testing the limits of wrapping sheets}
\label{sec:anaylsis}
So far, we have shown that the idea of wrapping a mesh sheet onto an image achieves strong performance across all benchmarks. But one might wonder, how good is a planar sheet prior for novel view synthesis in any arbitrary images? To test the limits of wrapping a mesh \textit{sheet}, we test it over a large variety of images including outdoor scenes, outdoor objects, indoor scenes, indoor objects, and even artistic paintings. We analyze our underlying mesh data structure by pretraining depth~\cite{lasinger2019towards} to fill in the $z$ values, and examine how well it generates novel views in-the-wild. As shown in Figure~\ref{fig:vis_popup} (top row), although missing a few details (such as tree branches), a scene mesh sheet captures the geometric structures sufficient enough to render high-resolution (960$\times$960) photorealistic novel views even from very large viewpoint changes. Please see videos at \href{https://worldsheet.github.io}{\textcolor{black}{\texttt{worldsheet.github.io}}} for animation of continuously generated views. This result confirms that our mesh sheet data structure and warping procedure, despite being simple, are flexible enough to handle the variety of the visual world.

\subsection{Analysis: multi-layered Worldsheets}
\label{sec:exp_multiple_layers}
As described in Sec.~\ref{sec:method_multiple_layers}, to better handle sharp depth discontinuities, we explore the extension to stack multiple mesh sheet layers, so that foreground objects and the background can be placed on different layers.
Interestingly, we observe that this extension does not make a noticeable improvement in RealEstate10K evaluation metrics compared to a single sheet (26.62 vs. 26.74 in PSNR) suggesting that single Worldsheet is sufficient enough for view synthesis application.
However, qualitatively speaking, we notice that it allows better handling of occlusions and parallax effect in our model under large viewpoint changes, as shown in Figure~\ref{fig:multiple_layers}. Please see supplemental for more details.

\begin{figure}[t]
\centering
\includegraphics[width=\linewidth]{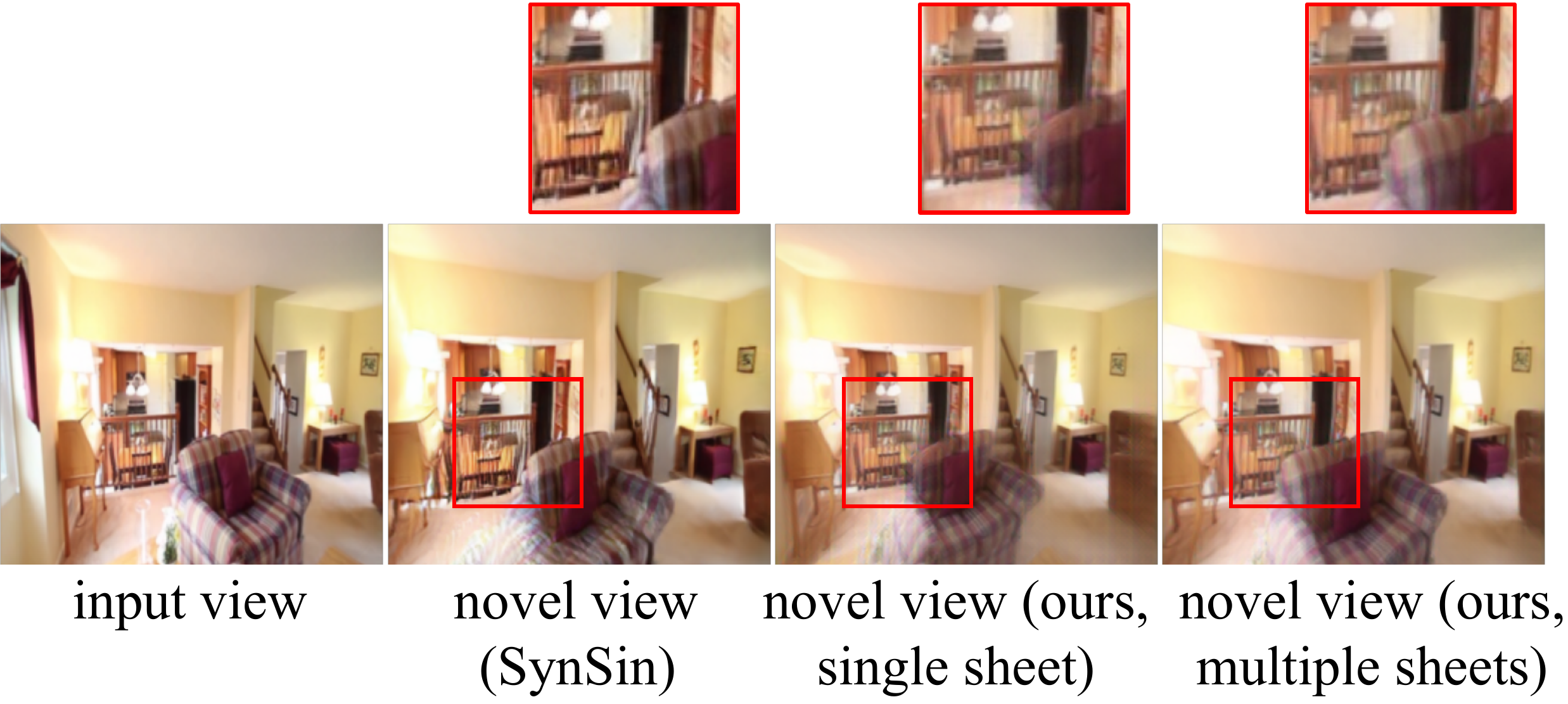} 
\vspace{-2em}
\caption{Using multiple mesh sheets (Sec.~\ref{sec:exp_multiple_layers}), our model (rightmost) better handles occlusions and generates parallax effect (\eg sofa occluding the handrail when moving camera to the right) compared to SynSin \cite{wiles2020synsin} or a single sheet.}
\label{fig:multiple_layers}
\vspace{-1.5em}
\end{figure}

\section{Conclusion}
In this work, we propose \methodname, which synthesizes novel views from a single image by shrink-wrapping the scene with a grid mesh. Our approach jointly learns the scene geometry and generates novel views through differentiable texture sampling and mesh rendering, supervised with only 2D images of the input and the target views. The approach is category-agnostic and end-to-end trainable, resulting in state-of-the-art performance on single-image view synthesis across three datasets by a large margin. 
\small{\noindent \textbf{Acknowledgments.} We are grateful to Alyosha Efros, Angjoo Kanazawa, Shubham Goel, Devi Parikh, Ross Girshick, Georgia Gkioxari, Justin Johnson, Brian Okorn and other colleagues at FAIR and CMU for fruitful discussions. This work was supported in part by DARPA Machine Common Sense grant (associated with D. Pathak and not associated with Facebook Inc).}

{\small
\bibliographystyle{ieee_fullname}
\bibliography{egbib}
}

\appendix
\normalsize

\counterwithin{figure}{section}
\counterwithin{table}{section}
\counterwithin{equation}{section}

\twocolumn[{
\begin{center}
\Large 
\textbf{\methodname: Wrapping the World in a 3D Sheet\\for View Synthesis from a Single Image}\\(Supplementary Material)
\par
\end{center}
\vspace{2em}
}]

\section{Continuous and large viewpoint changes}

Our approach allows synthesizing continuous novel views by smoothly moving to a new camera pose that is largely different from the input. We kindly request the readers to view the videos at \href{https://worldsheet.github.io}{\textcolor{black}{\texttt{worldsheet.github.io}}} to better understand the performance of our method. In these videos, we compare our synthesized novel views (from a single image) to SynSin \cite{wiles2020synsin} on the RealEstate10K dataset with simulated large view-point changes (the first frame contains the input view and the rest of the frames are synthesized). From the videos, it can be seen that our model can generate novel views with much larger camera translation and rotation than in the training data, while SynSin often suffers from severe artifacts in these cases, likely because its refinement network does not generalize well to a sparser point cloud (resulting from large camera zoom-in or rotation).

We also show in these videos continuously synthesized novel views on high resolution ($960 \times 960$) images over a wide range of scenes (the first frame is the input view), as described in our analysis in Sec.~4.3 in the main paper.

\section{Ablation study: using depth supervision}
\label{sec:supp_depth_supervision}

\begin{table*}
\footnotesize
\begin{center}
\setcounter{magicrownumbers}{0}
\begin{tabular}{clcccccccccccc}
\toprule
& &\multicolumn{9}{c}{Matterport \cite{chang2017matterport3d}} & \multicolumn{3}{c}{Replica \cite{straub2019replica}} \\
 \cmidrule(lr){3-11} \cmidrule(lr){12-14}
& & \multicolumn{3}{c}{PSNR $\uparrow$} & \multicolumn{3}{c}{SSIM $\uparrow$} & \multicolumn{3}{c}{Perc Sim $\downarrow$} & PSNR $\uparrow$ & SSIM $\uparrow$ & Perc Sim $\downarrow$ \\ 
\# & Method & Both & InVis & Vis & Both & InVis & Vis & Both & InVis & Vis & \\ 
\cmidrule(lr){1-2} \cmidrule(lr){3-5} \cmidrule(lr){6-8} \cmidrule(lr){9-11} \cmidrule(lr){12-14}
\rownumber & Im2Im \cite{zhou2016view} & 15.87 & 16.20 & 15.97 & 0.53 & 0.60 & 0.48 & 2.99 & 0.58 & 2.05 & 17.42 & 0.66 & 2.29 \\
\rownumber & Tatarchenko \etal \cite{tatarchenko2016multi} & 14.79 & 14.83 & 15.05 & 0.57 & 0.62 & 0.53 & 3.73 & 0.74 & 2.50 & 14.36 & 0.68 & 3.36 \\
\rownumber & Vox \cite{sitzmann2019deepvoxels} w/ UNet & 18.52  & 17.85 & 19.05 & 0.57 & 0.57 & 0.57 & 2.98 & 0.77  & 1.96 & 18.69 & 0.71 & 2.68 \\
\rownumber & Vox \cite{sitzmann2019deepvoxels} w/ ResNet & 20.62 & 19.64 & 21.22 & 0.70 & 0.69 & 0.68 & 1.97 & 0.47 & 1.19 & 19.77 & 0.75 & 2.24 \\
\midrule
\rownumber & SynSin \cite{wiles2020synsin} & 20.91 & 19.80 & 21.62 & 0.71 & 0.71 & 0.70 & 1.68 & 0.43 & 0.99 & 21.94 & 0.81 & 1.55 \\
\rownumber & SynSin (w/ depth sup.) \cite{wiles2020synsin} & \demph{21.59} & \demph{20.32} & \demph{22.46} & \demph{0.72} & \demph{0.71} & \demph{0.71} & \demph{1.60} & \demph{0.43} & \demph{0.92} & \demph{22.54} & \demph{0.80} & \demph{1.55} \\
\midrule
7 & ours & \textbf{24.67} & \textbf{22.90} & \textbf{26.00} & \textbf{0.82} & \textbf{0.77} & \textbf{0.82} & \textbf{1.05} & \textbf{0.35} & \textbf{0.54} & \textbf{23.51} & \textbf{0.85} & \textbf{1.32} \\
8 & ours (w/ depth sup.) & \demph{24.75} & \demph{22.85} & \demph{26.18} & \demph{0.82} & \demph{0.77} & \demph{0.82} & \demph{1.06} & \demph{0.36} & \demph{0.54} & \demph{22.78} & \demph{0.84} & \demph{1.51} \\
\bottomrule
\end{tabular}
\end{center}
\vspace{-1.5em}
\caption{Analyses of our model and SynSin using explicit depth supervision during training (line 8 and 6) on the Matterport dataset. Our model without depth (the default setting; line 7) performs almost equally as well as its counter part with depth supervision (line 8) on Matterport and generalizes better to Replica, outperforming both variants of SynSin (line 5 and 6). See Sec.~\ref{sec:supp_depth_supervision} for details.}
\label{tab:supp_exp_mp3d_replica_depth_loss}
\vspace{-1em}
\end{table*}

In our experiments in the paper, we show that our model can be trained using only two views of a scene \textit{without} 3D or depth supervision. In this section, we further analyze our approach by training it \textit{with} depth supervision on the Matterport dataset, where the ground-truth depth can be obtained from the Habitat simulator.

In this analysis, we modify the differentiable mesh renderer to render RGB-D images from our mesh, and apply an L1 loss between the ground-truth and the rendered depth as additional supervision. We also compare with the performance of SynSin with depth supervision (reported in \cite{wiles2020synsin}). The results are shown in Table~\ref{tab:supp_exp_mp3d_replica_depth_loss}. It can be seen that our model without depth supervision (the default setting; line 7) works almost equally as well as its counterpart using depth supervision (line 8) on the Matterport dataset and generalizes better to the Replica dataset. In addition, it outperforms SynSin under both supervision settings (lines 5-6).

\section{Additional analyses on RealEstate10K}
\label{sec:supp_additional_analyses_realestate10k}

\begin{table*}[t]
\footnotesize
\begin{center}
\setcounter{magicrownumbers}{0}
\begin{tabular}{clccccccccc}
\toprule
& &\multicolumn{9}{c}{RealEstate10K \cite{zhou2018stereo} (averaged metrics over all predictions)} \\
 \cmidrule(lr){3-11}
& & \multicolumn{3}{c}{PSNR $\uparrow$} & \multicolumn{3}{c}{SSIM $\uparrow$} & \multicolumn{3}{c}{Perc Sim $\downarrow$} \\ 
\# & Method & Both & Peripheral & Center & Both & Peripheral & Center & Both & Peripheral & Center \\ 
\cmidrule(lr){1-2} \cmidrule(lr){3-5} \cmidrule(lr){6-8} \cmidrule(lr){9-11}
\rownumber & Im2Im \cite{zhou2016view} & 15.56 & 15.65 & 15.80 & 0.51 & 0.53 & 0.44 & 2.59 & 1.93 & 0.65 \\
\rownumber & Tatarchenko \etal \cite{tatarchenko2016multi} & 11.11 & 11.32 & 10.68 & 0.32 & 0.35 & 0.24 & 3.96 & 2.96 & 1.13 \\
\rownumber & SynSin~\cite{wiles2020synsin} & 20.47 & 20.35 & 21.65 & 0.68 & 0.68 & 0.68 & 1.49 & 1.16 & 0.29 \\
\rownumber & SynSin w/ R-50 backbone\dag & 19.37 & 19.33 & 20.18 & 0.65 & 0.65 & 0.64 & 1.64 & 1.26 & 0.34 \\
\rownumber & Single-View MPI~\cite{tucker2020single} & 21.17 & 20.90 & 23.03 & 0.70 & 0.70 & 0.71 & 1.59 & 1.27 & 0.31 \\
\midrule
\rownumber & ours ($33\times 33$ mesh) & 23.11 & 22.90 & 24.83 & \textbf{0.75} & 0.74 & \textbf{0.76} & 1.17 & 0.94 & \textbf{0.21} \\
\rownumber & ours ($65\times 65$ mesh) & \textbf{23.41} & \textbf{23.20} & \textbf{25.17} & \textbf{0.75} & \textbf{0.75} & \textbf{0.76} & \textbf{1.14} & \textbf{0.92} & \textbf{0.21} \\
\bottomrule
\end{tabular}
\end{center}
\vspace{-1.5em}
\caption{Performance of our and previous approaches on the RealEstate10K dataset with metrics averaged over all predictions, which is consistent with our evaluation on Matterport and Replica in Sec. 4.1 in the main paper. We evaluate on the entire $256 \times 256$ image (\textbf{both}), the \textbf{center} $128 \times 128$ crop (nearly always visible), and the remaining \textbf{peripheral} regions (containing most of the invisible regions). See Sec.~\ref{sec:supp_additional_analyses_realestate10k} for details. (\dag: We also evaluate a variant of SynSin by replacing its U-Net backbone with the same ResNet-50 backbone pretrained on ImageNet as used in our model.)}
\label{tab:supp_realestate10k_details}
\vspace{-1em}
\end{table*}

As described in Sec. 4.2 in the main paper, we follow the evaluation protocol of SynSin \cite{wiles2020synsin} on the RealEstate10K dataset. To enable comparison with StereoMag \cite{zhou2018stereo} that uses two input views on this dataset, in \cite{wiles2020synsin}, the best metrics of two views were reported for single-view methods. At test time, for each target view, this involves making two separate predictions based on two different input views respectively, and then selecting the best metrics between the two predictions as the score for this target view. Note that this evaluation protocol is only applied to the RealEstate10K dataset (in Table 3 and 4 in the main paper) and is not applied to Matterport or Replica.

In this section, we further evaluate by taking the average metrics over all predictions to measure how well the model does on average from a single input view and to be consistent with our evaluation on Matterport and Replica in Table 1 and 2 in the main paper. Apart from metrics over the entire image, we would also like to analyze how well each model does on rendering regions seen in the input view vs. invisible regions (where things must be imagined). However, we cannot compute the exact visibility map as there are no ground-truth geometry annotations in the RealEstate10K dataset. To get an approximation, we evaluate on the central $\frac{W_{im}}{2} \times \frac{H_{im}}{2} = 128 \times 128$ crop of the target image (\textbf{Center}, which is nearly always visible) and the rest of the image (\textbf{Peripheral}, containing most of the invisible regions).

The results of these analyses are shown in Table~\ref{tab:supp_realestate10k_details}. It can be seen that our method achieves the highest performance on both the center regions (which are mostly visible) and the peripheral regions, outperforms the previous single-view based approaches by a large margin.

Our model uses a simple ResNet-50 backbone with an output stride of 8 pixels to extract image features (see Sec. 3.1 in the main paper), while SynSin \cite{wiles2020synsin} adopts a U-Net backbone that has a higher output feature resolution same as the input image (\ie output feature stride is 1 pixel). To further study the impact of different backbone architectures, we train a variant of SynSin by replacing its U-Net backbone with the same ResNet-50 backbone pretrained on ImageNet (and upsampling its output feature map to stride 1 with a deconvolution layer) to be consistent with our model, shown in line 4 in Table~\ref{tab:supp_realestate10k_details}. Comparing it with line 3 or 6, it can be seen that this variant of SynSin with ResNet-50 backbone performs worse than the default SynSin architecture, or our model. This suggests that SynSin requires a high-resolution feature output to build a per-pixel point-cloud for view synthesis, while our model is able to work with a lower resolution (a larger stride) in the backbone.

\section{Details on differentiable texture sampler}

Our differentiable texture sampler (Sec.~3.2 in the main paper) splats image pixels onto the texture map through each face. This splatting procedure involves three main steps: forward-mapping, normalization, and hole filling.

Suppose the flow $(u,v)=f(i, j)$ maps image coordinates $(i, j)$ to UV coordinates $(u,v)$ on the texture map, which can be obtained through mesh rasterization (here we drop the face index $k$ for simplicity). To implement $\hat{T} = \mathrm{splat}(I_{in}, f)$ (Eqn.~5 in the main paper), we first forward-map the image pixels to the texture map, using bilinear assignment when the mapped texture coordinates $(u,v)$ do not fall on integers, as $\mathrm{forward\_map}$ below.
\begin{verbatim}
function T = forward_map(I, f)
  T = all_zeros(W_uv, H_uv)
  for i in 1:W_im
    for j in 1:H_im
      u, v = f(i, j)
      uf, uc = floor(u), ceil(u)
      vf, vc = floor(v), ceil(v)
      T[uf,vf] += I[i,j] * (uc - u) * (vc - v)
      T[uc,vf] += I[i,j] * (u - uf) * (vc - v)
      T[uf,vc] += I[i,j] * (uc - u) * (v - vf)
      T[uc,vc] += I[i,j] * (u - uf) * (v - vf)
    end
  end
  return T
\end{verbatim}
In the implementation above, gradients can be taken over $f$ through the bilinear weights.

However, forward mapping alone will lead to incorrect pixel intensity (\eg imagine down-scaling an image to half its width and height by forward-mapping -- each pixel in the low-resolution image will receive assignment from 4 pixels and become $4\times$ brighter). Hence, a second normalization step is applied:
\begin{eqnarray}
\hat{T}_{sum} &=& \mathrm{forward\_map}(I_{in}, f) \\
\hat{W}_{sum} &=& \mathrm{forward\_map}(I_{one}, f) \\
\hat{T}_{norm} &=& \hat{T}_{sum} / \max\left(\hat{W}_{sum}, 10^{-4}\right)
\end{eqnarray}
where $I_{one}$ is an $W_{im}\times H_{im}$ image with all ones as its pixel intensity. $\hat{T}_{norm}$ contains the normalized splatting result. A threshold $10^{-4}$ is applied to avoid division by zero (which could happen due to holes described below).

\begin{figure}[t]
\centering
\includegraphics[width=\linewidth]{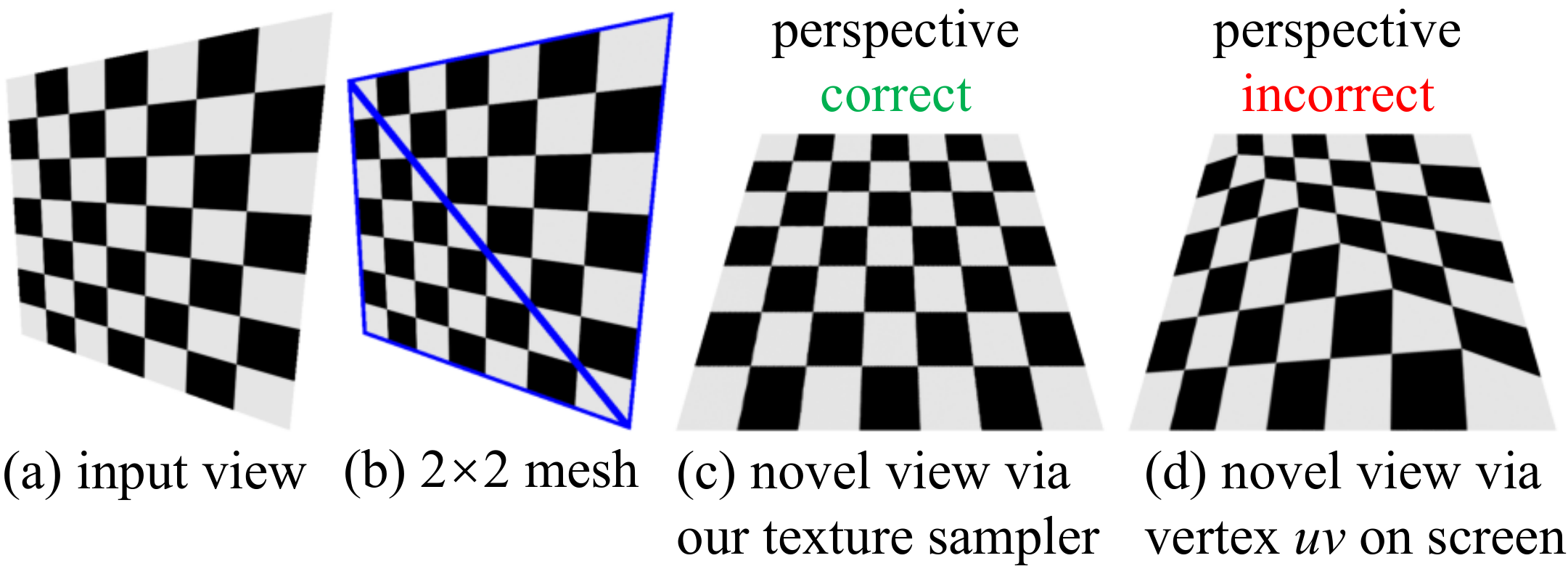}
\vspace{-1.5em}
\caption{Our texture sampler is perspective-correct.}
\label{fig:supp_texture_sampler}
\vspace{-1em}
\end{figure}

\myparagraph{Filling holes with Gaussian filtering} It is well known that the bilinear forward mapping above often leads to holes in the output (\eg imagine up-scaling an image to a much larger size -- there will be gaps in the output image as some pixels will not receive assignments). To minimize hole occurrence, in our UV texture map we assign the UV coordinates of each mesh vertex with an equally-spaced $W_m \times H_m$ lattice grid, and use the same image size as the texture map size ($W_{uv} \times H_{uv} = W_{im} \times H_{im}$). This ensures that most texels on the texture map receive assignments in forward mapping, so that holes rarely occur in $\hat{T}_{norm}$. However, to address corner cases, we further apply a Gaussian filter to fill the holes in $\hat{T}_{norm}$ (where $\hat{W}_{sum}$ is zero as no assignment is received from forward-mapping):
\begin{eqnarray}
\hat{M} &=& \mathrm{I}\left[\hat{W}_{sum} > 0\right] \\
\hat{T}_{g} &=& \left(F_g * \hat{T}_{norm}\right) / \left(F_g * \hat{M}\right) \\
\hat{T} &=& \hat{M} \cdot \hat{T}_{norm} + \left(\left(1 - \hat{M}\right) \cdot \hat{T}_{g}\right) \label{eqn:supp_texture_out}
\end{eqnarray}
where $F_g$ is a discrete 2D Gaussian kernel for image filtering (we use kernel size 7 and standard deviation 2 for $F_g$ in our implementation). Here $\hat{M}$ is a binary mask indicating which pixels have received assignments in forward mapping (\ie 1 means valid and 0 means holes), $\hat{T}_{g}$ is the Gaussian-blurred version of $\hat{T}_{norm}$ (where the division ensures the correct pixel intensity; otherwise it will be darker due to holes in $\hat{T}_{norm}$) and is used to fill only the holes in $\hat{T}_{norm}$. We use $\hat{T}$ in Eqn.~\ref{eqn:supp_texture_out} as the final splatting output.

\myparagraph{Perspective correctness} A main purpose of our differentiable texture sampler is to build perspective-correct novel views during texture reconstruction. We note that the alternative solution of directly using the input image as a texture map by putting vertex $uv$ texture coordinates in the input screen space for mesh rendering breaks perspective correctness, as shown in Figure~\ref{fig:supp_texture_sampler} (d). For perspective-correct novel views, one needs to invert the texture-map-to-image perspective transform when building UV texture maps from the image, which we implement in our texture sampler shown in Figure~\ref{fig:supp_texture_sampler} (c).

\section{Details on multi-layered Worldsheets}
\label{sec:supp_multple_layers}

\begin{figure}[t]
\small
\centering
\includegraphics[width=.85\linewidth]{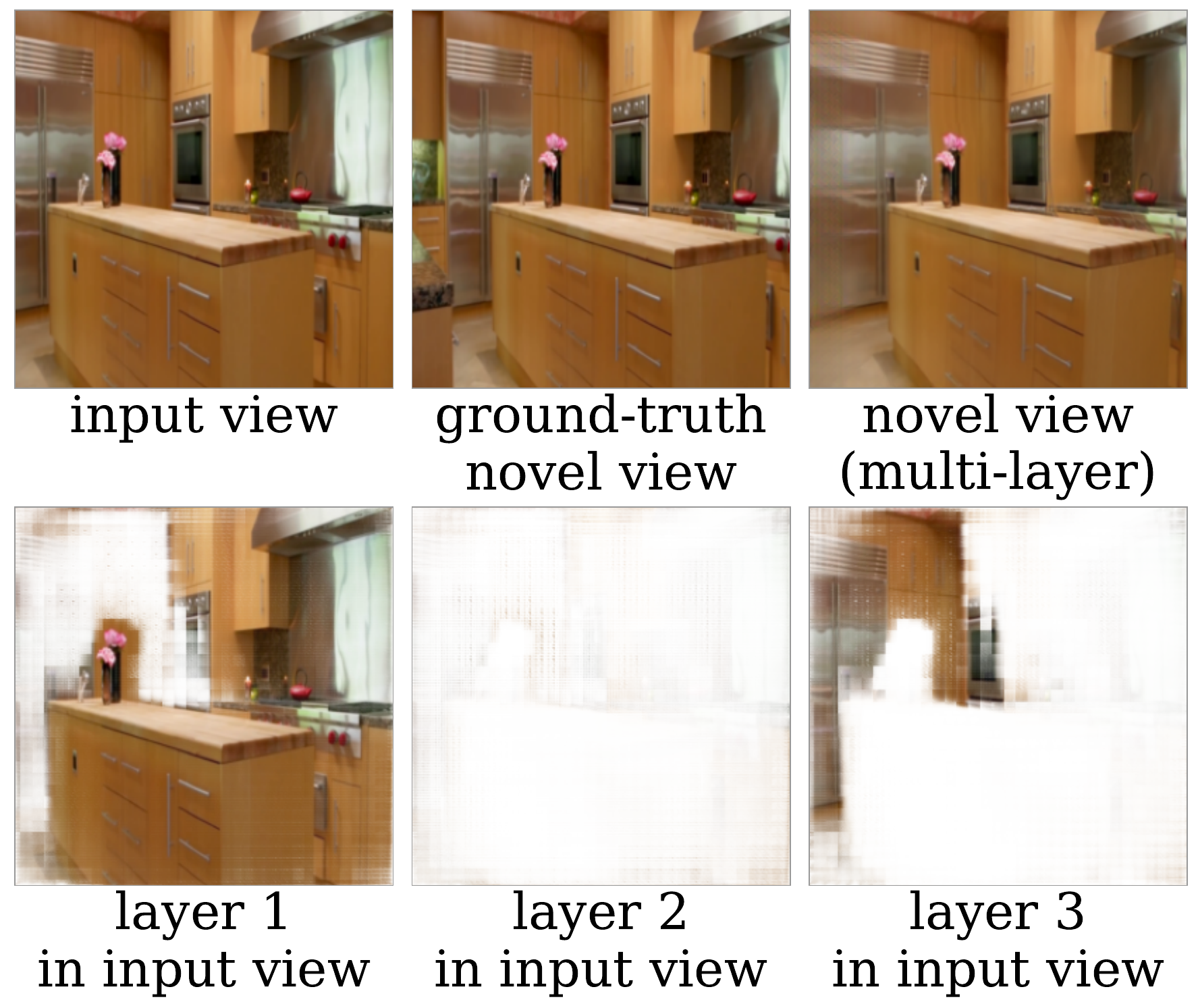}
\includegraphics[width=.85\linewidth]{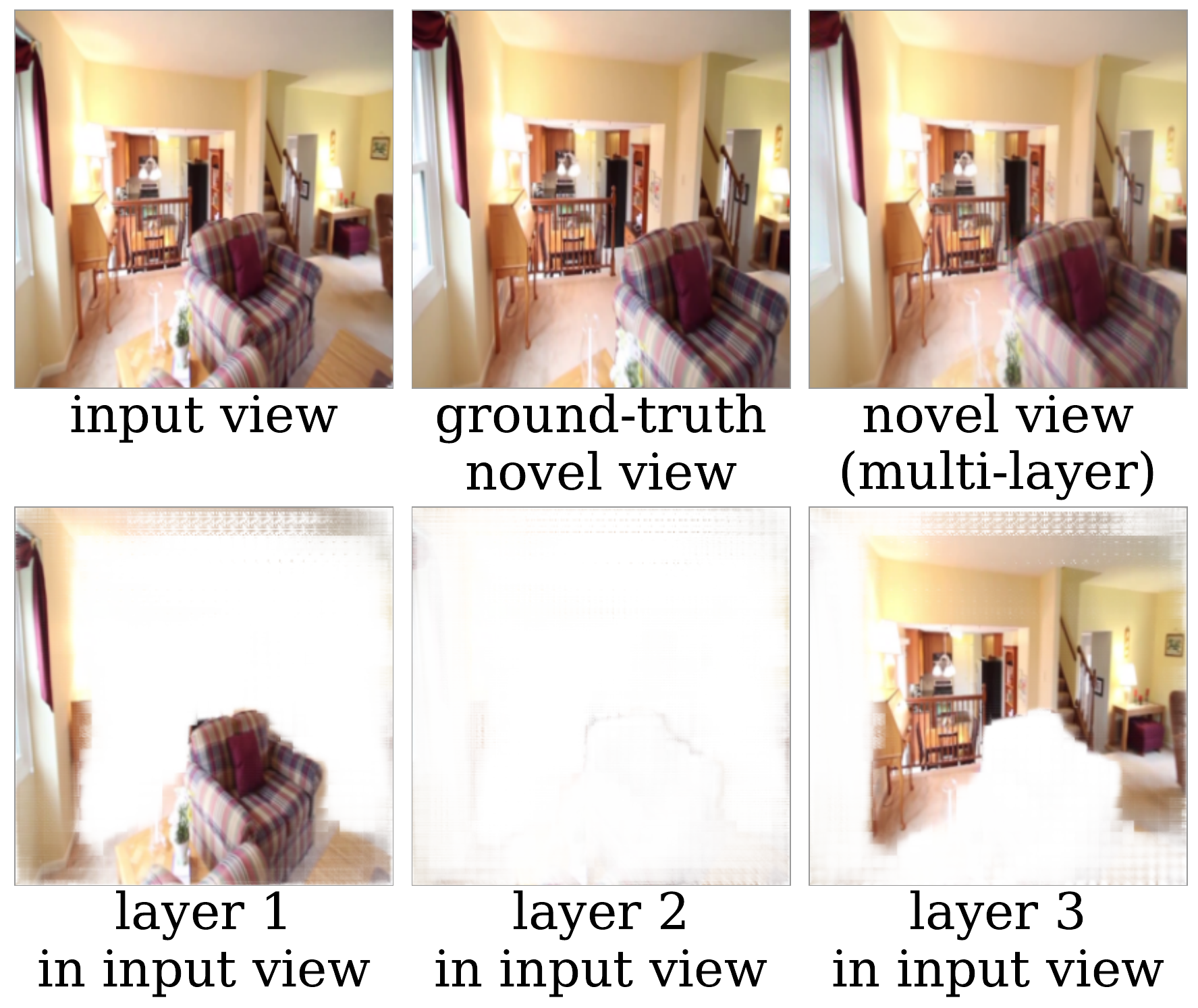}
\includegraphics[width=.85\linewidth]{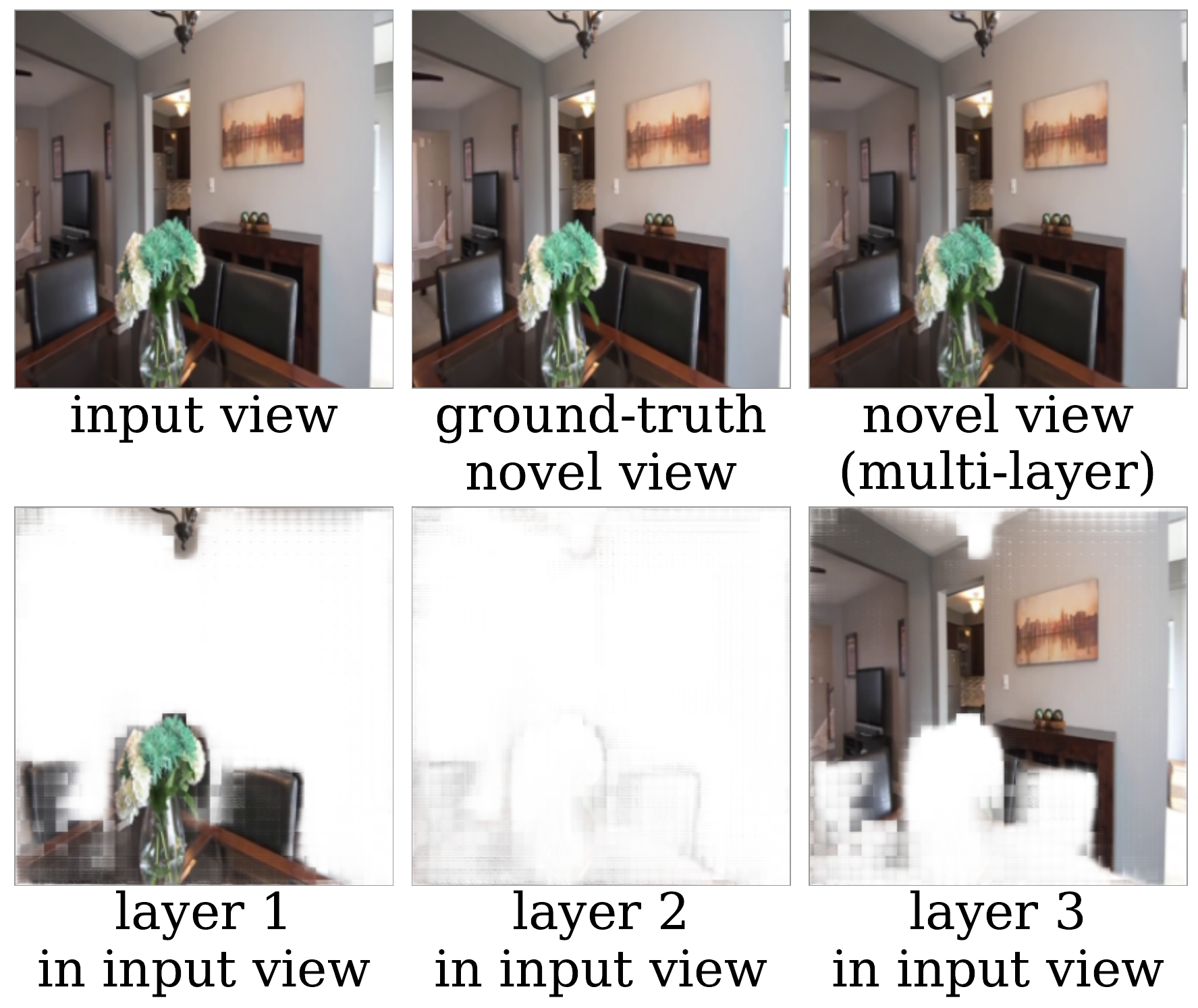}
\caption{View synthesis examples on RealEstate10K from our extension on multi-layered Worldsheets (see Sec.~\ref{sec:supp_multple_layers} for details). The top rows show the predicted novel views, while the bottom rows visualize each mesh layer $l=1,...,L$ (here $L=3$) along with its RGBA texture map in the input view (white is transparent). Through end-to-end training with only 2D rendering losses, the model learns to place scene structures at different depth levels onto different mesh layers, and separate foreground objects (\eg kitchen counter, sofa, or table) from their background.}
\label{fig:supp_multiple_layers}
\vspace{-3em}
\end{figure}

In our proposed Worldsheet model, we build a scene mesh by warping a planar sheet onto the scene. This model is capable of handling moderate occlusion and generating plausible novel views by deforming the mesh along object boundaries and refining the predicted novel view with an inpainting network. However, we also acknowledge that artifacts can sometimes occur in occluded regions or object boundaries when the disparity is very large between the input view and the novel view. This is partly because our current approach of deforming a mesh onto the scene does not capture all the fine-grained geometric details (such as the flower boundary as shown in the last two failure cases in the supplemental videos. We believe there is room for improvement in this direction (\eg via adaptive resolution), which we are interested in exploring in future work.

In Sec. 3.5 in the main paper, we propose a simple extension with multi-layered Worldsheets. The main purpose of this extension is to separate objects or scene structures at different depth levels into different mesh layers, instead of placing them all on a single sheet. In this extension, the 3D mesh geometry of each layer provides the geometric support for the scene components, while the transparency channel in the RGBA texture maps allows segmentation between different components. For example, to represent a sofa object, a mesh layer can be wrapped onto a larger 3D surface region covering the sofa surface, with its texture map containing the sofa texture over the object region while being transparent on the surrounding regions.

Specifically, we predict and warp a total of $L$ mesh sheets (\ie $L$ layers) onto the scene for view synthesis. For each layer $l=1,\cdots,L$, we predict its grid offset $\left(\Delta \hat{x}_{w,h}^{(l)}, \Delta \hat{y}_{w,h}^{(l)}\right)$ and its depth $z_{w,h}^{(l)}$ from the convolutional feature map $\{q_{w,h}\}$ similar to Sec. 3.1, and also predict a pixel-wise transparency map $\alpha^{(l)}$ of size $H_{im} \times W_{im}$ in the screen space of the input view as follows.
\begin{eqnarray}
\Delta \hat{x}_{w,h}^{(l)} &=& \frac{\tanh\left(W_1^{(l)} q_{w,h} + b_1^{(l)} \right)}{W_{m} - 1} \label{eqn:multilayer_offset_x}\\
\Delta \hat{y}_{w,h}^{(l)} &=& \frac{\tanh\left(W_2^{(l)} q_{w,h} + b_2^{(l)} \right)}{H_{m} - 1} \label{eqn:multilayer_offset_y} \\
z_{w,h}^{(l)} &=& g\left(W_3^{(l)} q_{w,h} + b_3^{(l)}\right) \label{eqn:multilayer_depth}
\label{eqn:multilayer_offset_y} \\
\left\{\alpha_{i,j}^{(l)}\right\} &=& \sigma\left(\mathrm{deconv}\left(\left\{q_{w,h}\right\}; W_4^{(l)}, b_4^{(l)}\right)\right) \label{eqn:multilayer_alpha}
\end{eqnarray}
Here $\alpha_{i,j}^{(l)}$ is the scalar alpha (\ie transparency) value at image pixel $(i,j)$,  $\mathrm{deconv}$ is a deconvolution (\ie transposed convolution) layer over the feature map with an output size equal to the image size, and $\sigma(\cdot)$ is the sigmoid function to transform the alpha values to the range between 0 and 1.

For each layer $l=1,\cdots,L$, we construct a corresponding 3D mesh sheet $M^{(l)}$ following the procedure in Sec. 3.1 and also build its UV texture map $\hat{T}^{(l)}$ consisting of RGBA channels by concatenating the predicted transparency values $\alpha_{i,j}^{(l)}$ with the input image and splatting them onto the mesh texture space using our differentiable texture sampler in Sec. 3.2. Finally, we render all the mesh faces from all $L$ layers $M^{(1)},\cdots,M^{(L)}$ in the novel view along with their RGBA UV texture maps $\hat{T}^{(1)},\cdots,\hat{T}^{(L)}$ through alpha compositing. The whole model can be trained end-to-end under the same supervision using only 2D rendering losses.

We use a total of $L=3$ layers in our analyses. Figure~\ref{fig:supp_multiple_layers} visualizes this extension on multi-layered Worldsheets. It can be seen that through end-to-end training, the model learns to place scene structures at different depth levels onto different mesh layers and separate foreground objects (\eg kitchen counter, sofa, or table) from their background.
We qualitatively find that it better handles occlusions and parallax effect under large viewpoint changes, as shown in Sec. 4.4 in the main paper.

\section{Hyper-parameters in our model}

In our nonlinear function $g(\cdot)$ to scale the network prediction into depth values (in Eqn. 3 in the main paper), we use different output scales based on the depth range in each dataset. On Matterport and Replica, we use
\begin{equation}
    g(\psi) = 1 / (0.75 \cdot \sigma(\psi) + 0.01) - 1.
\end{equation}
On RealEstate10K (which has larger depth range), we double the output depth scale and use
\begin{equation}
    g(\psi) = 2 / (0.75 \cdot \sigma(\psi) + 0.01) - 2.
\end{equation}
However, we find that the performance of our model is quite insensitive to the hyper-parameters in $g(\cdot)$.

In our differentiable texture sampler and the mesh renderer, we mostly follow the hyper-parameters in PyTorch3D \cite{ravi2020accelerating}. We use $K=10$ faces per pixel and 1e-8 blur radius in mesh rasterization, 1e-4 sigma and 1e-4 gamma in softmax RGB blending, and background color filled with the mean RGB intensity on each dataset. On Matterport and Replica, the input views have 90-degree field-of-view. On RealEstate10K, we multiply the actual camera intrinsic matrix of each frame into its camera extrinsic $R$ and $T$ matrices, so that we can still use the same intrinsics and 90-degree field-of-view in the renderer. On high resolution images in the wild (Sec.~4.3 in the main paper), we assume 45-degree field-of-view.

We choose our mesh size $W_m$ and $H_m$ based on the image size. In our experiments on Matterport and Replica (Sec.~4.1 in the main paper), we use $256\times 256$ input image resolution following SynSin \cite{wiles2020synsin}, and use pixel stride 8 on the lattice grid sheet (from which our mesh is built), resulting in $W_m = H_m = 1 + 256 / 8 = 33$. On the RealEstate10K dataset (Sec.~4.2 in the main paper), we additionally experiment with pixel stride 4 on the grid sheet, giving $W_m = H_m = 1 + 256 / 4 = 65$. In our analysis on high resolution images in the wild (resized to have the image long side equal to $960$ and padded to $960\times 960$ square size for ease of rendering in PyTorch3D; Sec.~4.3 in the main paper), we use $W_m \times H_m = 129 \times 129$ mesh on the actual image regions (not including the padding regions).

\section{More visualized examples}

\begin{figure}[t]
\centering
\includegraphics[width=\linewidth]{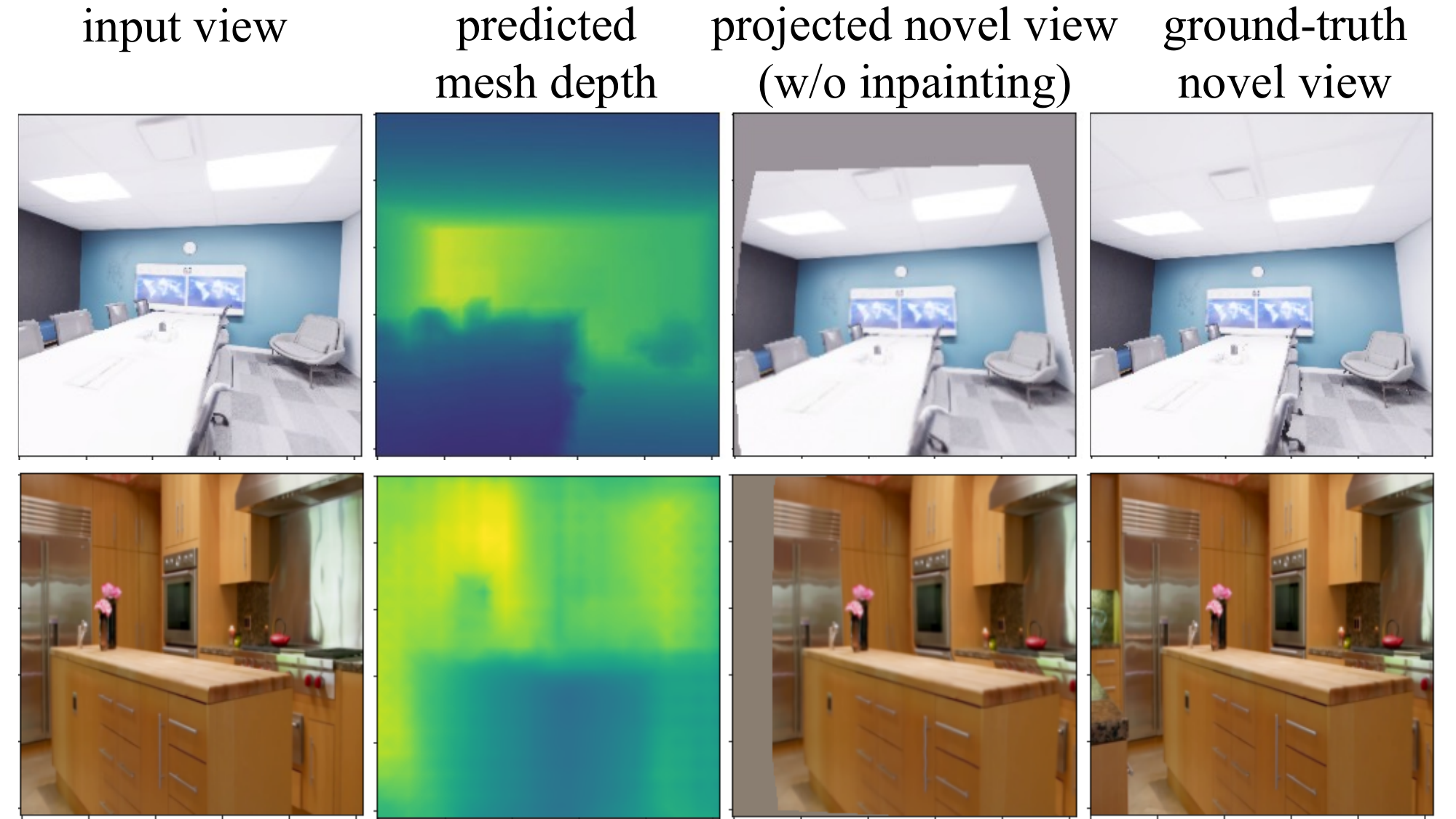}
\caption{Predicted depth maps from our scene mesh on Replica (upper) and RealEstate10K (lower).}
\label{fig:supp_depth_map}
\end{figure}

Figure~\ref{fig:supp_depth_map} shows the depth maps from our scene mesh, where most of the scene structure is captured, giving coherent novel view projections.

\begin{figure*}[t]
\vspace{3em}
\small
\centering
\includegraphics[width=0.48\textwidth]{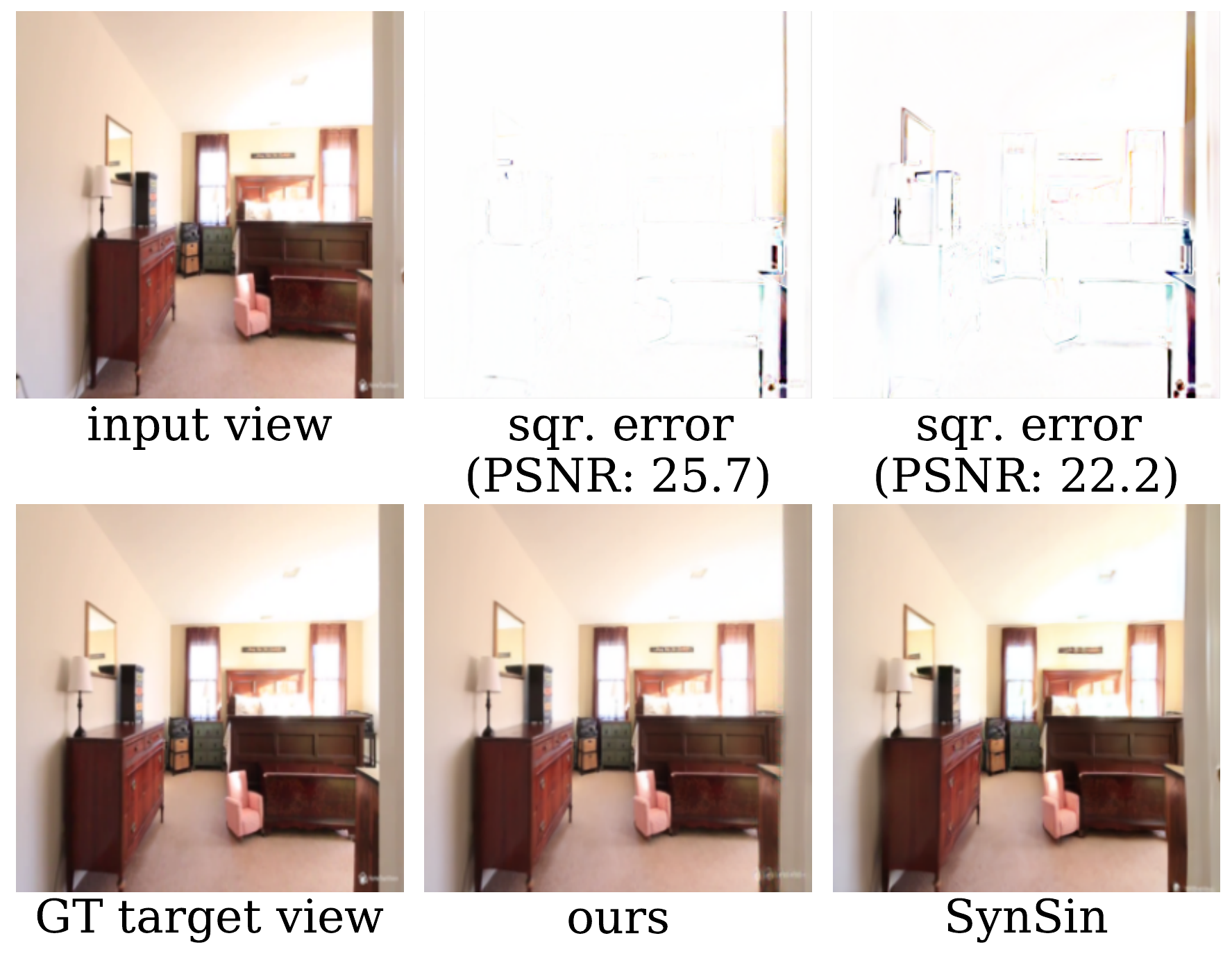}~~~~~~~~
\includegraphics[width=0.48\textwidth]{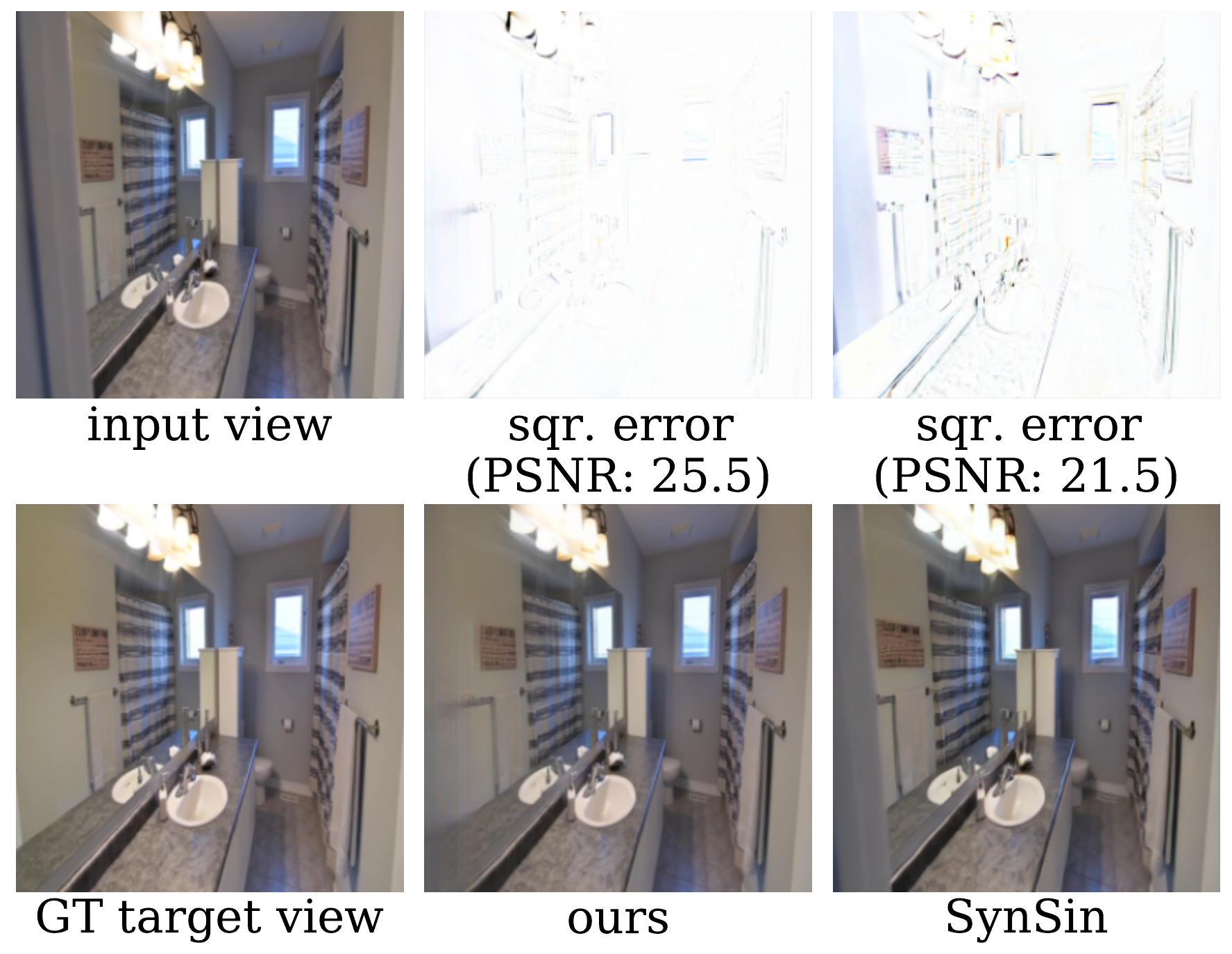} \\
\vspace{2em}
\includegraphics[width=0.48\textwidth]{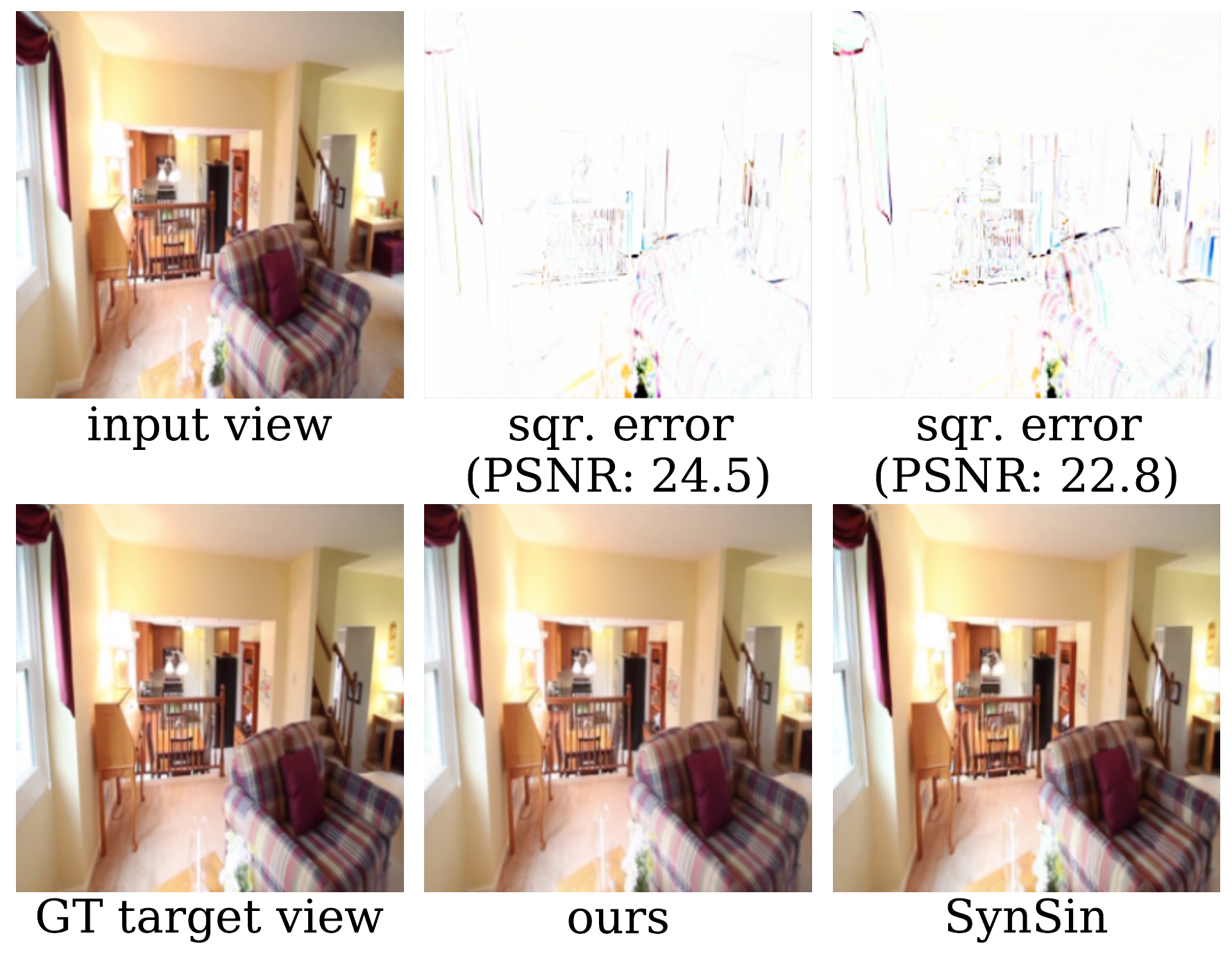}~~~~~~~~
\includegraphics[width=0.48\textwidth]{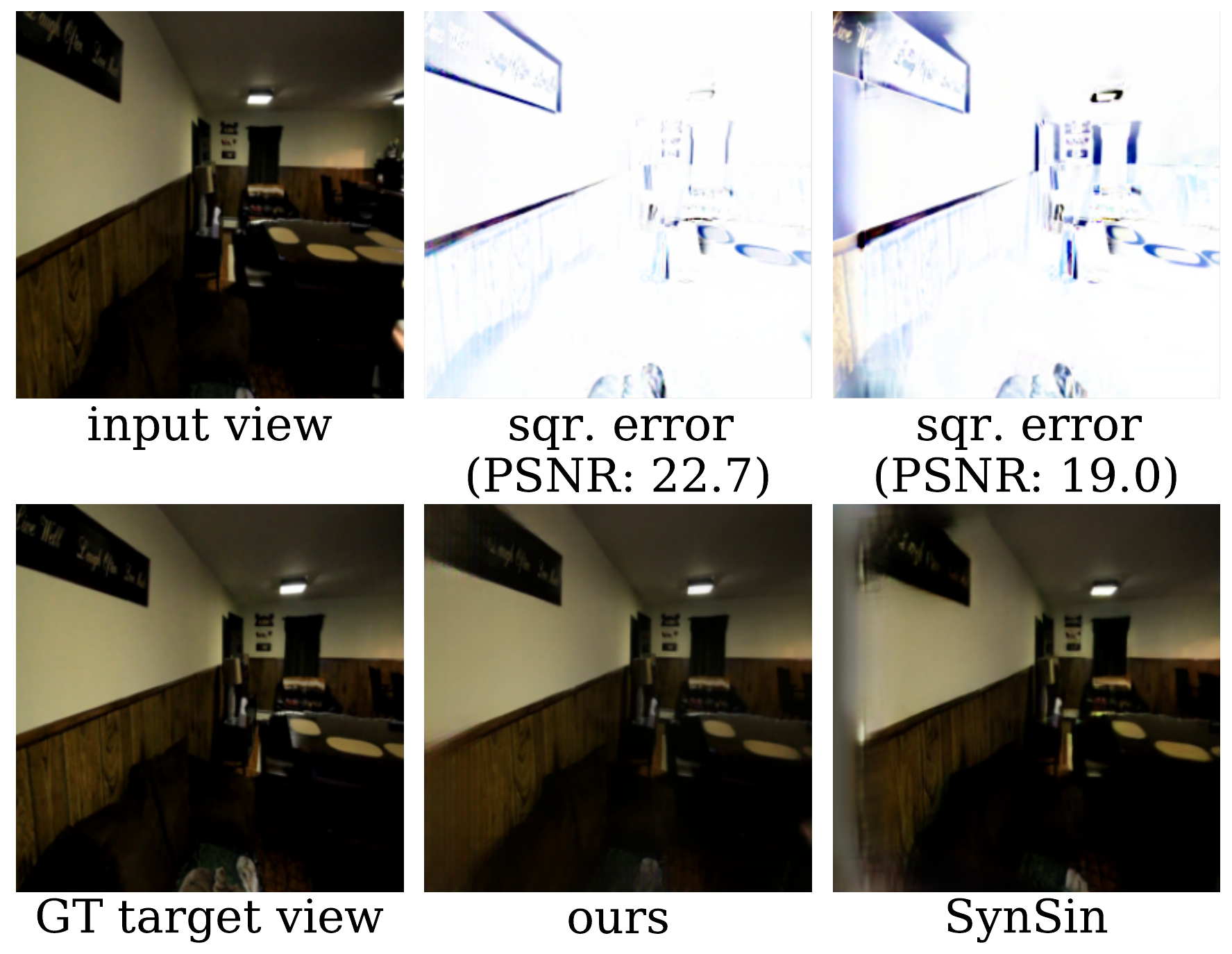} \\
\vspace{2em}
\includegraphics[width=0.48\textwidth]{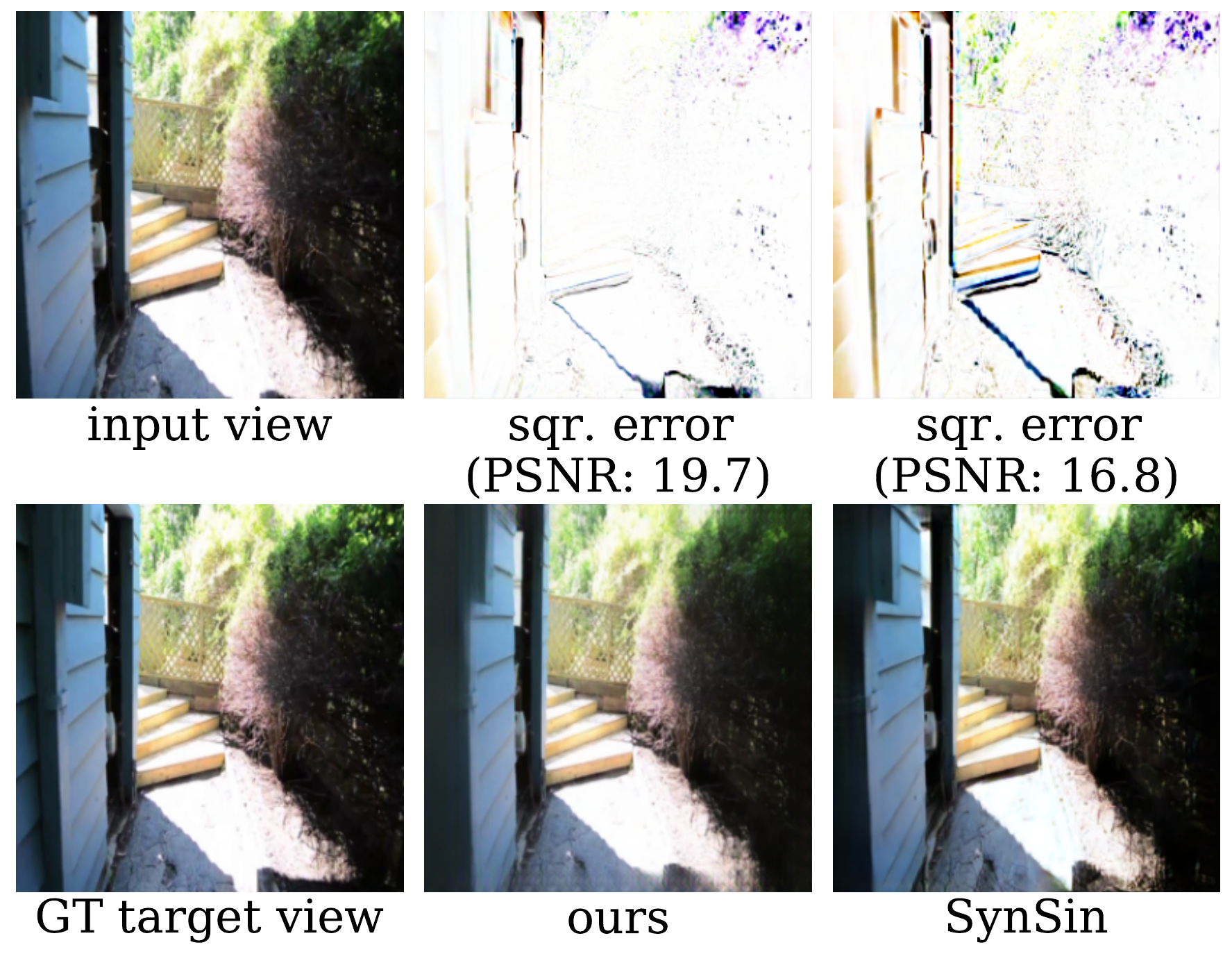}~~~~~~~~
\includegraphics[width=0.48\textwidth]{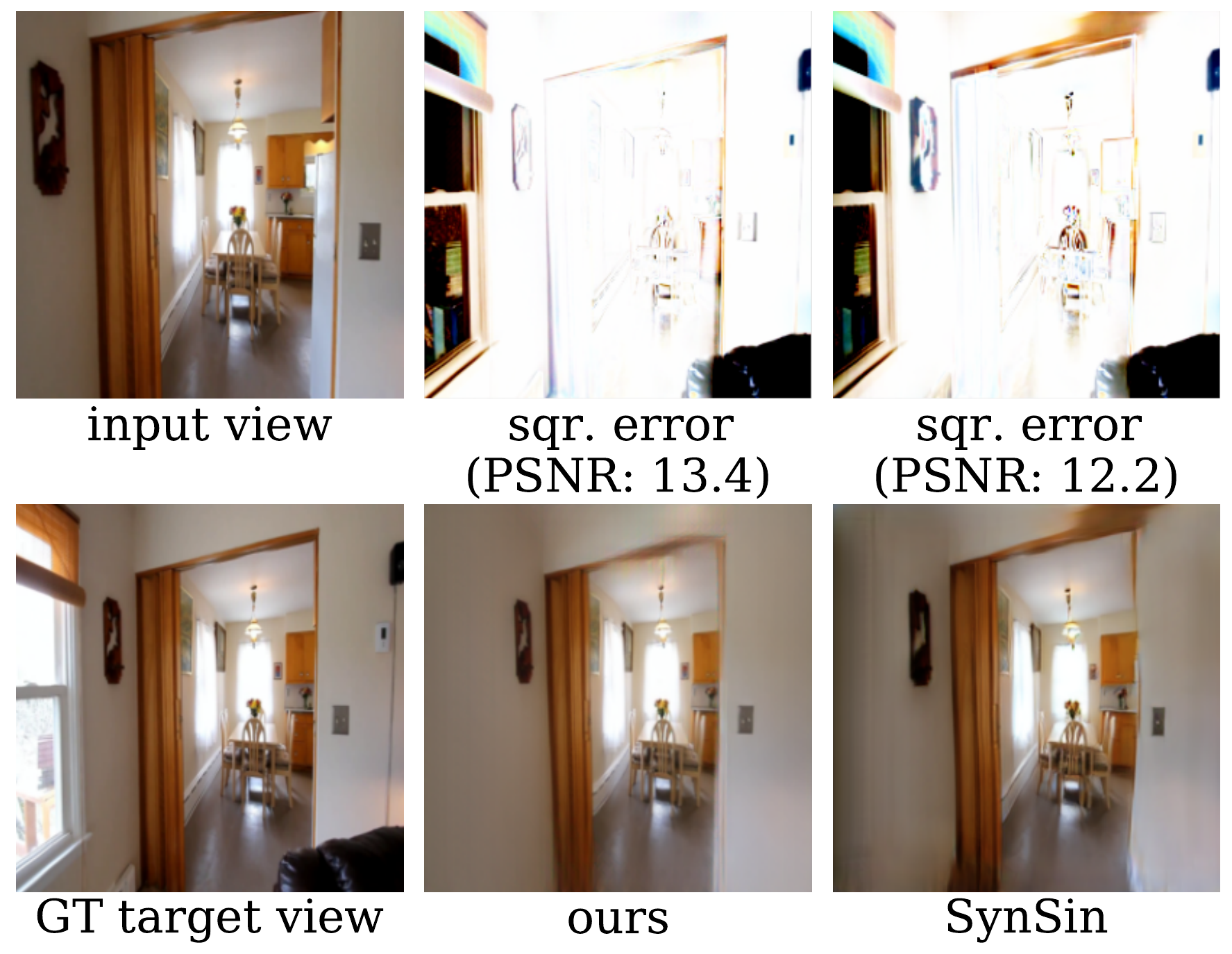} \\
\vspace{3em}
\end{figure*}

\begin{figure*}[t]
\small
\centering
\includegraphics[width=0.48\textwidth]{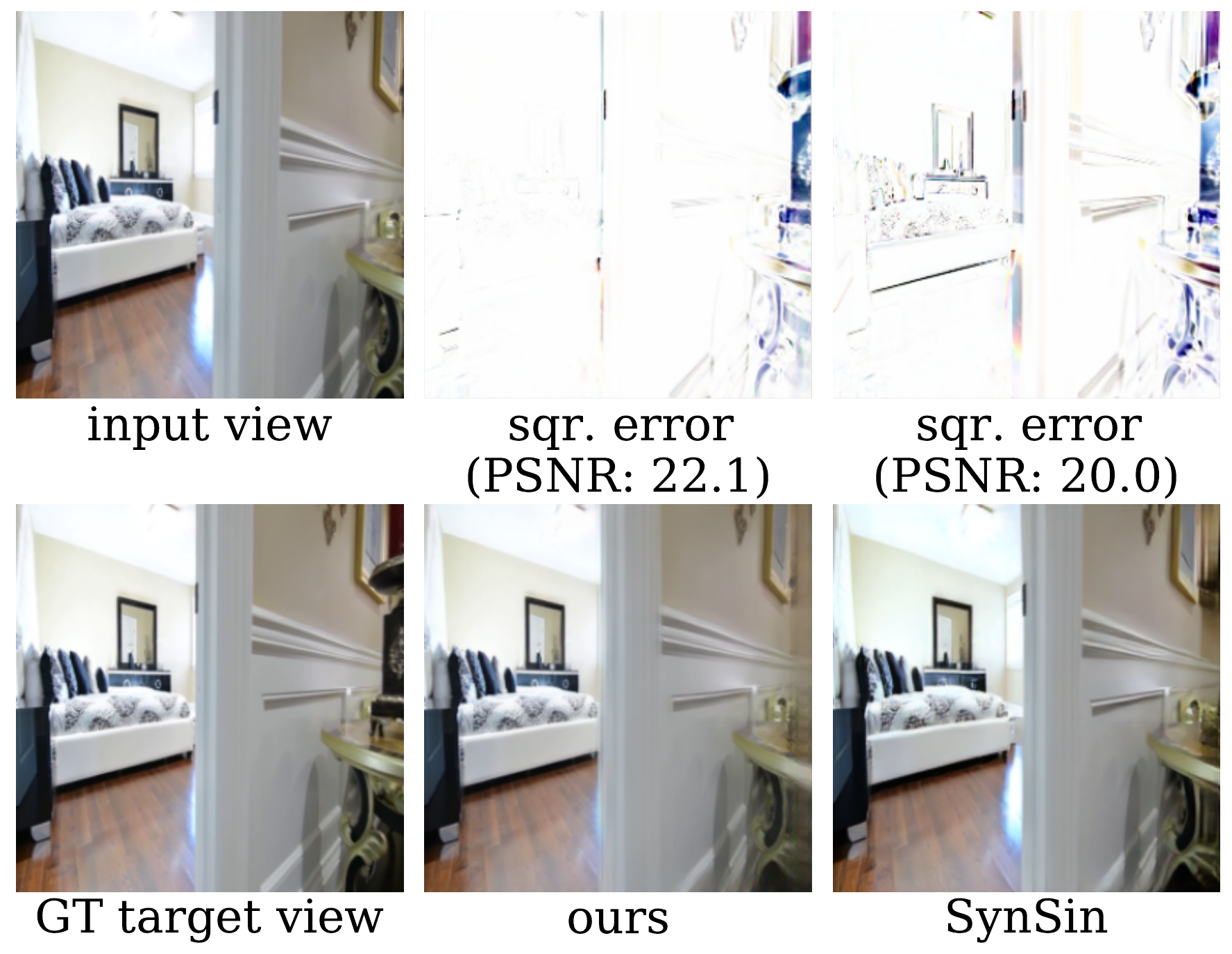}~~~~~~~~
\includegraphics[width=0.48\textwidth]{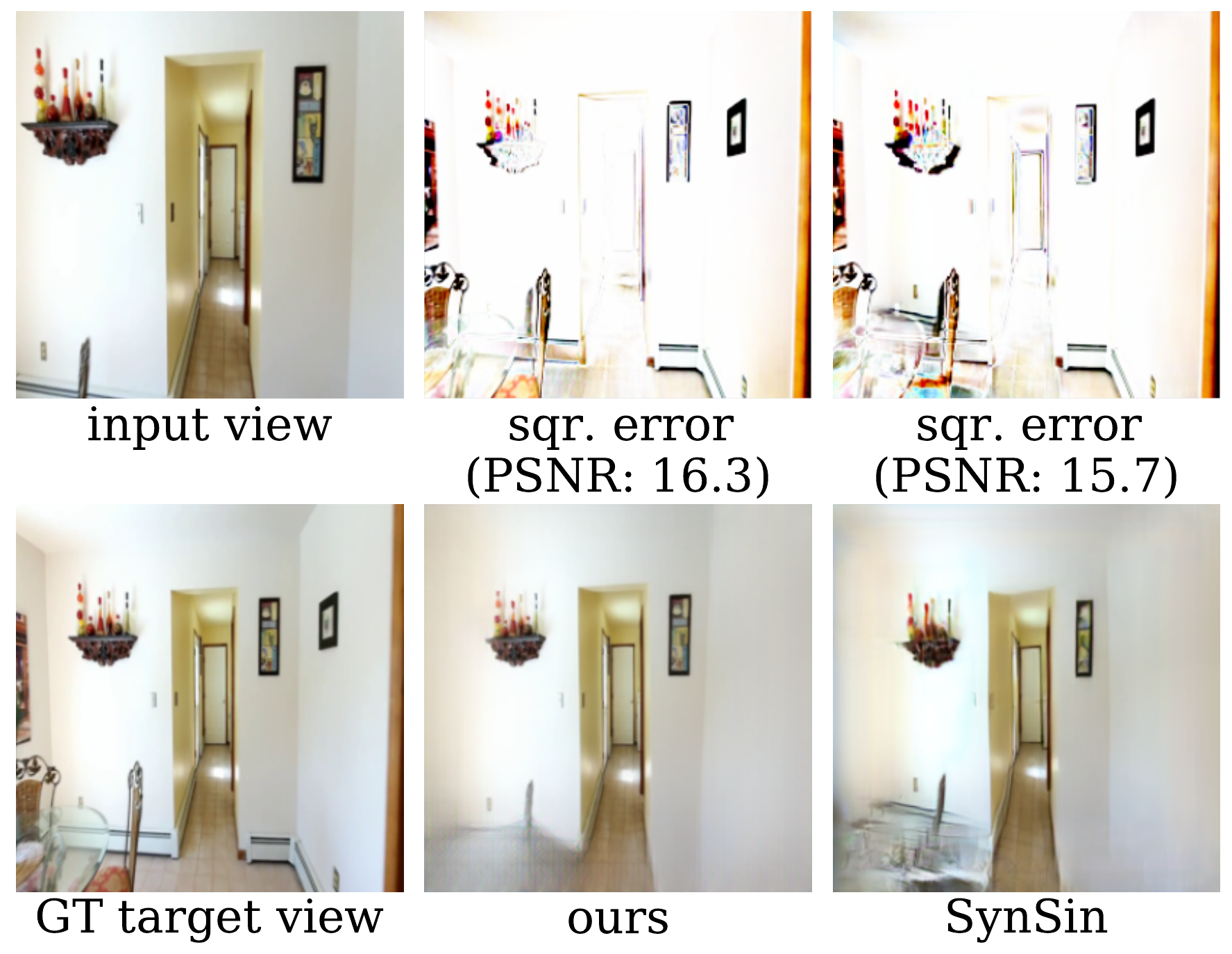} \\
\vspace{2em}
\includegraphics[width=0.48\textwidth]{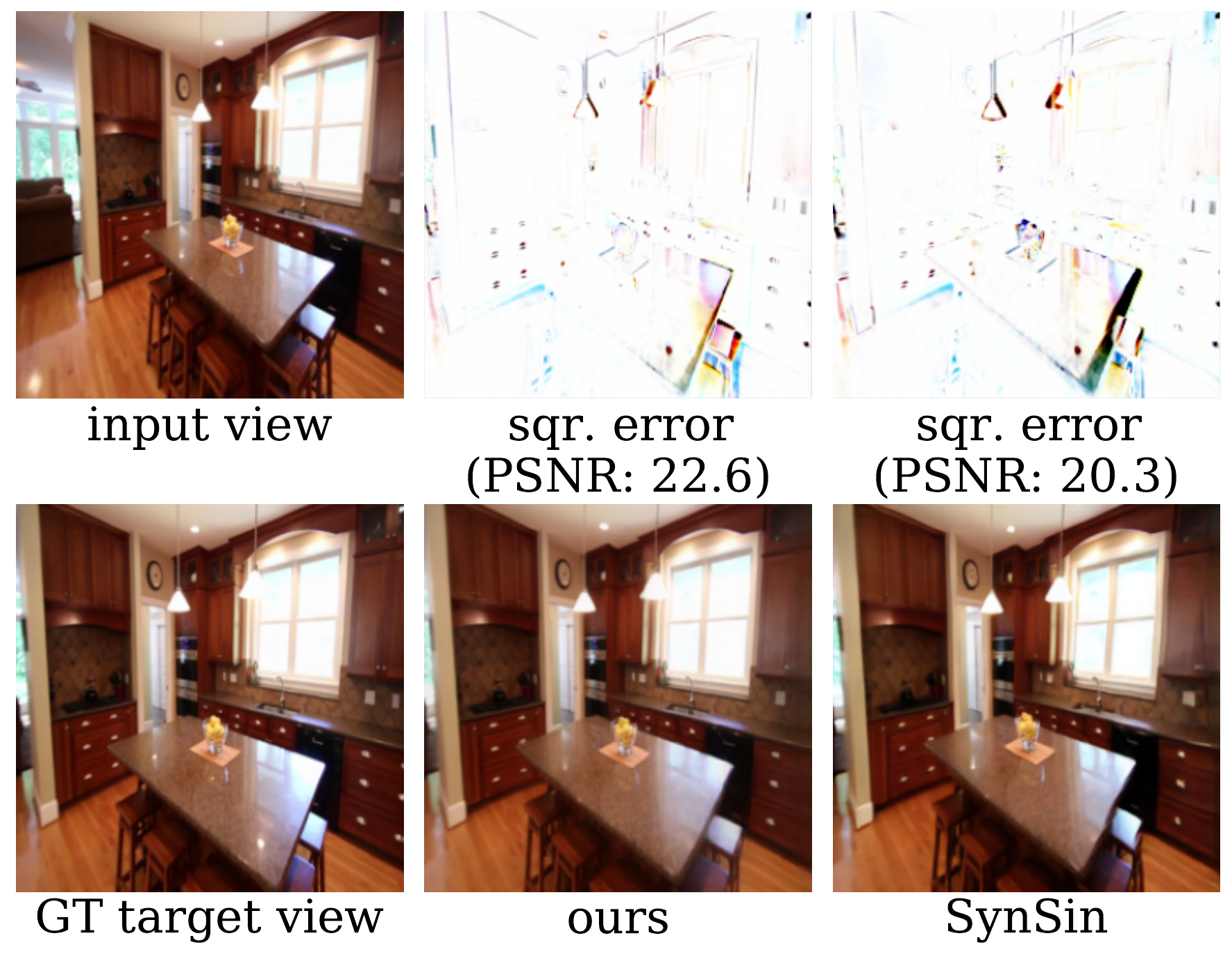}~~~~~~~~
\includegraphics[width=0.48\textwidth]{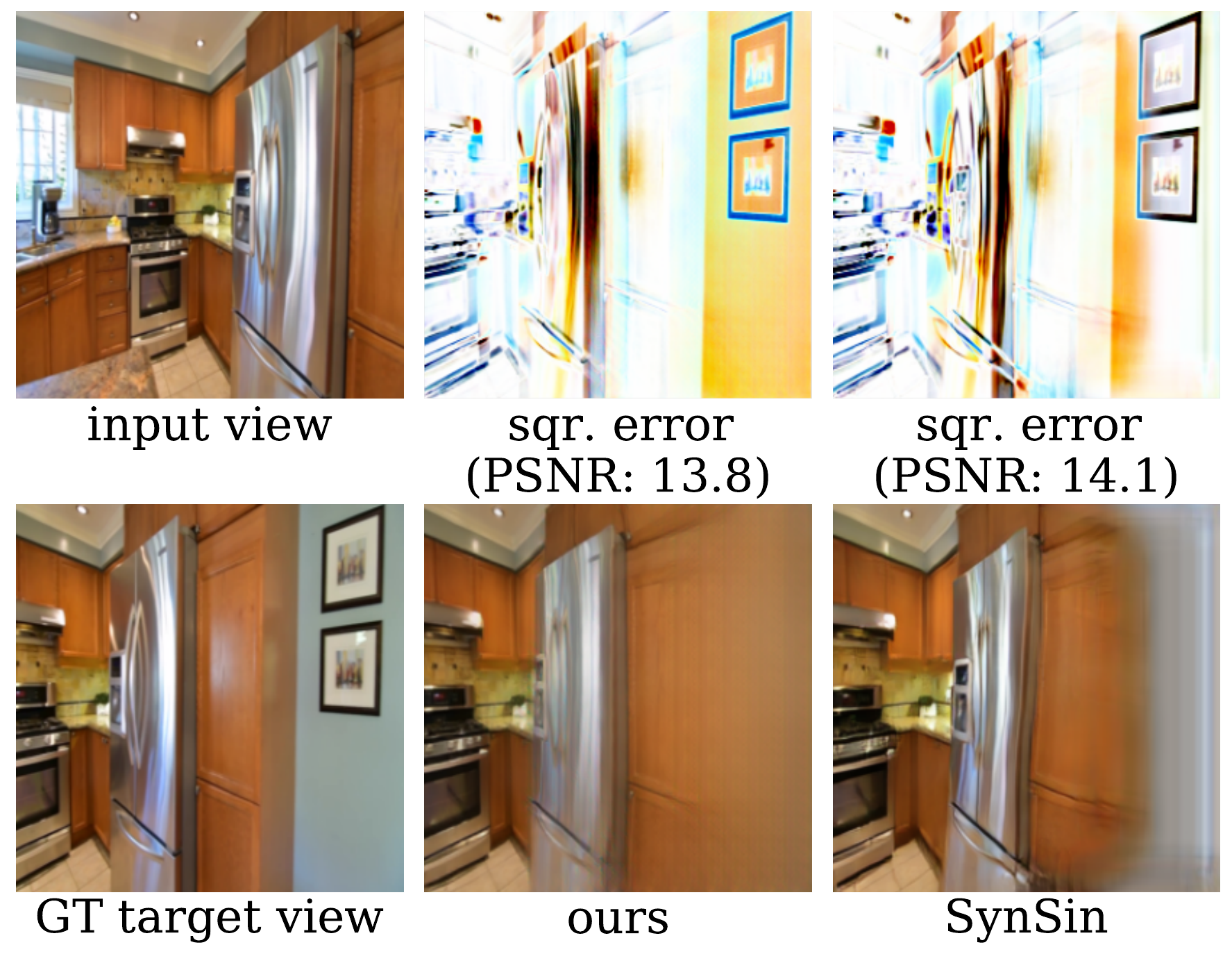} \\
\vspace{2em}
\includegraphics[width=0.48\textwidth]{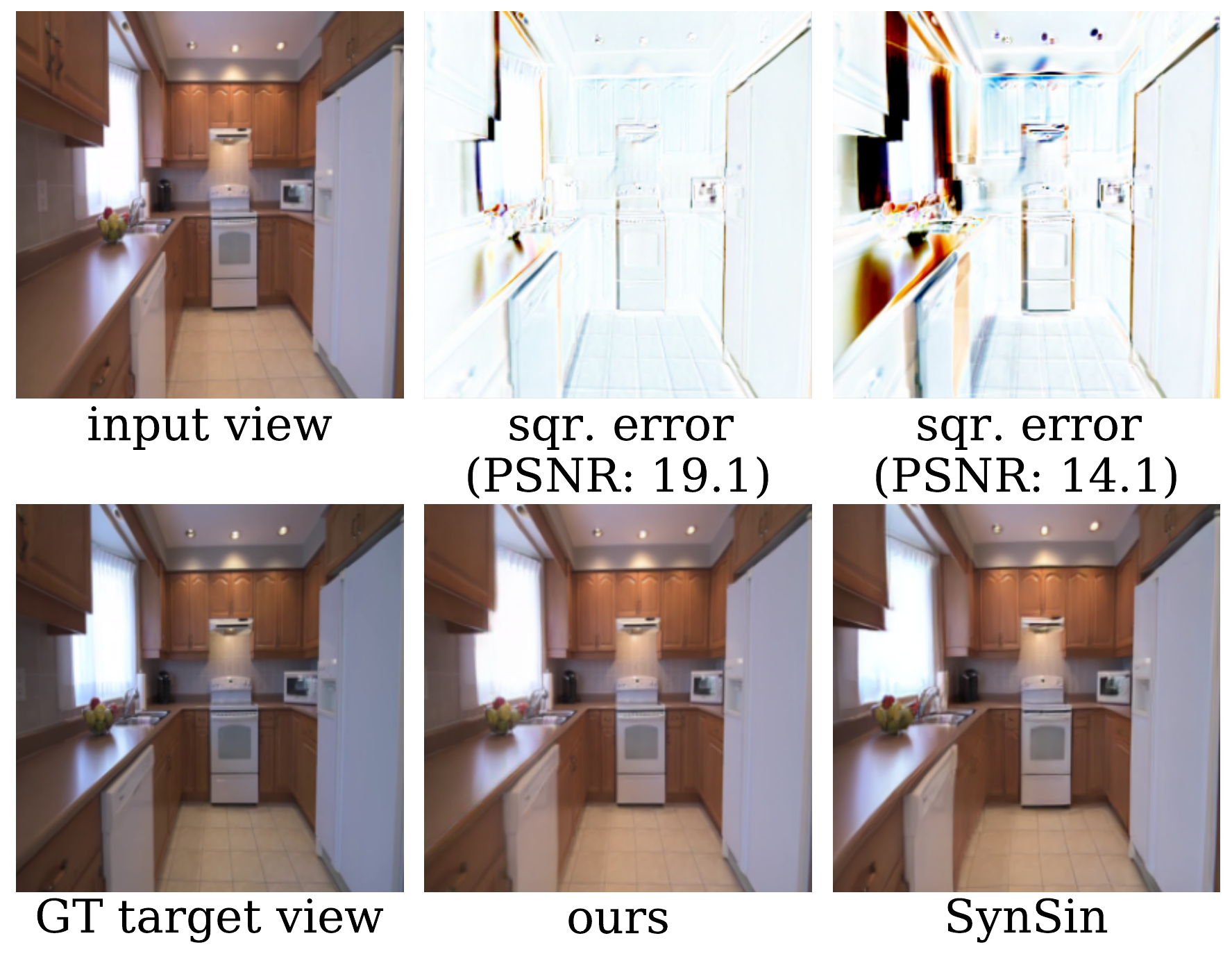}~~~~~~~~
\includegraphics[width=0.48\textwidth]{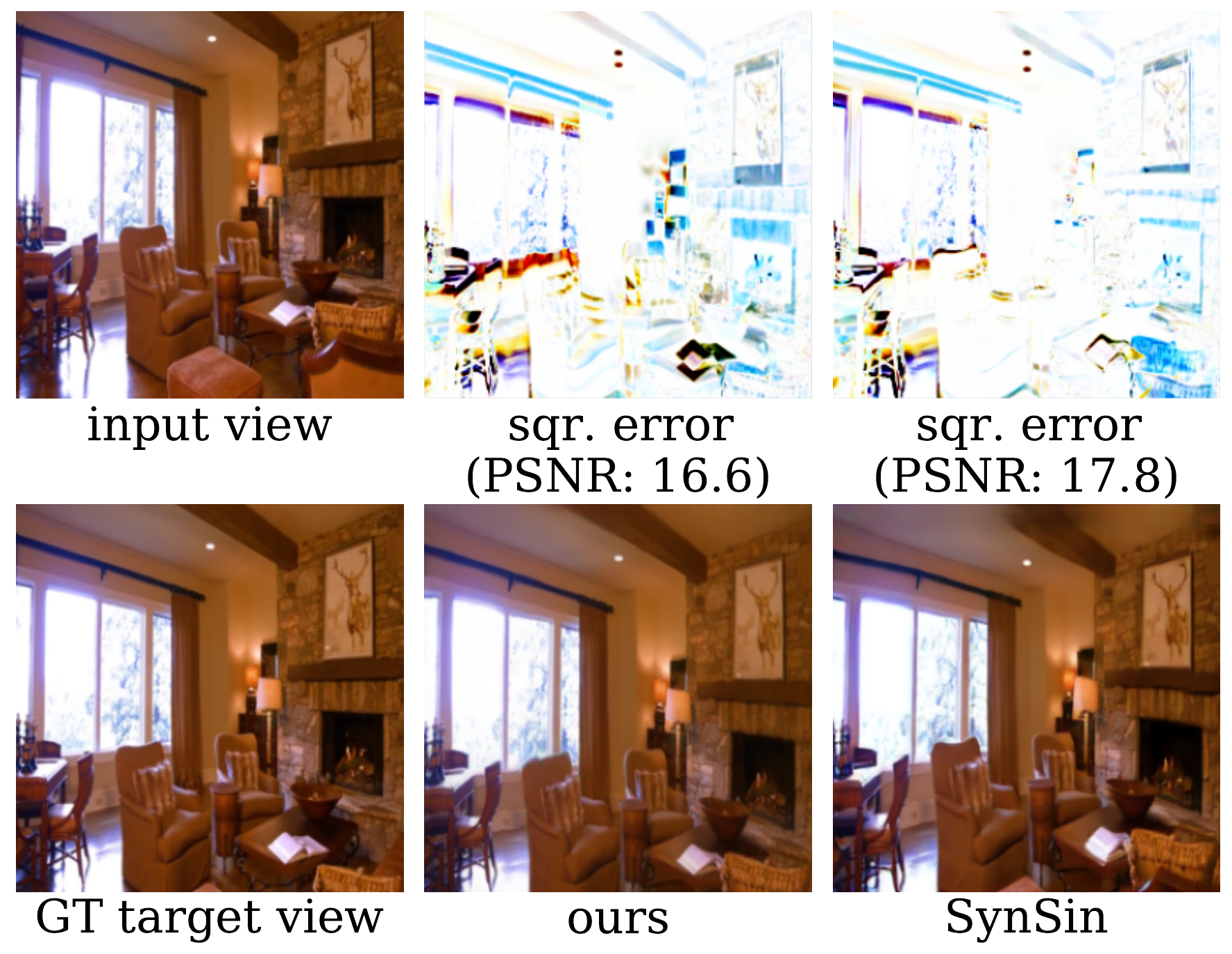} \\
\caption{Additional synthesized novel views (with squared error maps of the target view) from our method and SynSin \cite{wiles2020synsin} on the RealEstate10K dataset (darker is higher error). Our method paints things in the novel view at more precise locations, resulting in lower error and higher PSNR.}
\label{fig:supp_vis_realestate10k_psnr}
\end{figure*}

Figure~\ref{fig:supp_vis_realestate10k_psnr} shows additional visualization and error map comparisons between our approach and SynSin on the RealEstate10K dataset (similar to Figure 7 in the main paper), where our method paints things in the novel view at more precise locations with lower error and higher PSNR.

\end{document}